\documentclass{article}

\usepackage[dvipsnames, table]{xcolor}
\usepackage[preprint]{neurips_2025}

\usepackage[utf8]{inputenc} 
\usepackage[T1]{fontenc}    
\usepackage{hyperref}       
\usepackage{url}            
\usepackage{booktabs}       
\usepackage{amsfonts}       
\usepackage{nicefrac}       
\usepackage{microtype}      
\usepackage{adjustbox}

\usepackage{algorithm}
\usepackage[noend]{algpseudocode}
\usepackage{amsmath}
\usepackage{caption}
\usepackage{cleveref}
\usepackage{enumitem}
\usepackage{float}
\usepackage{graphicx}
\usepackage{makecell}
\usepackage{multirow}
\usepackage{natbib}
\usepackage{wrapfig}
\usepackage{pifont}
\usepackage{times}
\usepackage{tcolorbox}
\usepackage{latexsym}
\usepackage{tikz}
\usepackage{wrapfig}
\usepackage{tcolorbox}

\definecolor{maroon}{cmyk}{0,0.87,0.68,0.32}
\definecolor{myyellow}{RGB}{218, 160, 109}
\definecolor{brickred}{rgb}{0.8, 0.25, 0.33}
\definecolor{brandeisblue}{rgb}{0.0, 0.44, 1.0}
\definecolor{applegreen}{rgb}{0.55, 0.71, 0.0}
\definecolor{aogreen}{rgb}{0.0, 0.5, 0.0}

\definecolor{gdmb}{RGB}{47, 114, 173}  
\definecolor{gdmr}{RGB}{199, 100,  38}
\definecolor{gdmg}{RGB}{70, 155, 118}
\definecolor{gdmm}{RGB}{193, 126, 165}
\definecolor{gdmy}{RGB}{239, 227,  98}
\definecolor{gdmc}{RGB}{110, 179, 228}
\definecolor{gdmk}{RGB}{20, 20, 20}
\definecolor{turquoise}{cmyk}{0.65,0,0.1,0.3}
\definecolor{purple}{rgb}{0.65,0,0.65}
\definecolor{dark_green}{rgb}{0, 0.5, 0}
\definecolor{orange}{rgb}{0.8, 0.6, 0.2}
\definecolor{red}{rgb}{0.8, 0.2, 0.2}
\definecolor{darkred}{rgb}{0.6, 0.1, 0.05}
\definecolor{blueish}{rgb}{0.0, 0.3, .6}
\definecolor{light_gray}{rgb}{0.7, 0.7, .7}
\definecolor{pink}{rgb}{1, 0, 1}
\definecolor{greyblue}{rgb}{0.25, 0.25, 1}
\definecolor{orgred}{rgb}{1.0, 0, 0}

\usepackage{pifont}

\definecolor{sh_gray}{rgb}{0.84,0.84,0.84}
\definecolor{sh_gray2}{rgb}{1,0.89,0.75}
\definecolor{color3}{rgb}{0.95,0.95,0.95}
\definecolor{color4}{rgb}{0.94,0.94,1}
\definecolor{color5}{rgb}{1,0.96,0.88}

\definecolor{shuopurple}{rgb}{0.65,0,0.65}

\title{Demystifying the Visual Quality Paradox in Multimodal Large Language Models}
\author{
Shuo Xing$^1$\thanks{\  Equal Contribution.} \ \thanks{\ Email: \texttt{\{shuoxing,tzz\}@tamu.edu}} \ ,\quad
Lanqing Guo$^2$\footnotemark[1] \ ,\quad
Hongyuan Hua$^3$\footnotemark[1] \ ,\quad
Seoyoung Lee$^2$,\quad 
Peiran Li$^1$,\quad \\
\bf Yufei Wang$^4$,\quad
Zhangyang Wang$^2$,\quad
Zhengzhong Tu$^1$\footnotemark[2] \ \thanks{\  Corresponding author.}\\
\\
$^1$Texas A\&M University \quad
$^2$University of Texas at Austin \\
$^3$University of Toronto \quad 
$^4$Nanyang Technological University
}

\begin{document}

\maketitle

\begin{abstract}

Recent Multimodal Large Language Models (MLLMs) excel on benchmark vision-language tasks, yet little is known about how input visual quality shapes their responses. \textit{Does higher perceptual quality of images already translate to better MLLM understanding?} We conduct the first systematic study spanning leading MLLMs and a suite of vision-language benchmarks, applying controlled degradations and stylistic shifts to each image. Surprisingly, we uncover a visual-quality paradox: \textit{model, task, and even individual-instance performance can improve when images deviate from human-perceived fidelity}. Off-the-shelf restoration pipelines fail to reconcile these idiosyncratic preferences. To close the gap, we introduce Visual-Quality Test-Time Tuning (VQ-TTT)—a lightweight adaptation module that: (1) inserts a learnable, low-rank kernel before the frozen vision encoder to modulate frequency content; and (2) fine-tunes only shallow vision-encoder layers via LoRA. VQ-TTT dynamically adjusts each input image in a single forward pass, aligning it with task-specific model preferences. Across the evaluated MLLMs and all datasets, VQ-TTT lifts significant average accuracy, with no external models, cached features, or extra training data. These findings redefine ``better'' visual inputs for MLLMs and highlight the need for adaptive, rather than universally ``clean'', imagery, in the new era of AI being the main data customer.
\end{abstract}

\section{Introduction}
Multimodal Large Language Models (MLLMs), which extend the capabilities of traditional language models to jointly process vision and language inputs, have recently achieved remarkable progress across a wide range of benchmarks~\citep{li2022blip, li2023blip2, liu2024llava,llavanext,llama3.2, Qwen-VL, Qwen2VL, lu2024deepseek, wu2024deepseek, bai2025qwen2, qwen2.5, beyer2024paligemma,abdin2024phi,abouelenin2025phi}. These models excel in tasks such as visual question answering~\citep{singh2019towards,lu2022scienqa,Fu2023mme,hudson2019gqa,lu2024mathvista,gurari2018vizwiz,yue2023mmmu,xing2025can,li2025safeflow}, captioning~\citep{chen2015microsoftcoco,agrawal2019nocaps}, and image-text retrieval~\citep{xing2025re,yang2023re, yasunaga2022retrieval,hu2025mrag}, often demonstrating impressive zero-shot and few-shot generalization. However, despite this progress, a critical and understudied question remains: 
\begin{tcolorbox}[before skip=2mm, after skip=0.0cm, boxsep=0.0cm, middle=0.0cm, top=0.1cm, bottom=0.1cm]
    \textit{\textbf{(Q)}
    How does the perceptual quality of input images affect the performance of MLLMs? How to ensure robust performance when input images suffer from diverse degradations?
    }
\end{tcolorbox}
\vspace*{0.2cm}

Intuitively, one might expect that cleaner, sharper, and more natural-looking images would lead to better model understanding, mirroring the preferences of human observers. Yet, in this work, we present a counterintuitive and surprising phenomenon—what we term the \textbf{visual-quality paradox}: for many MLLMs, performance on vision-language tasks can \textit{improve} when images deviate from conventional notions of visual fidelity. This paradox challenges a common assumption in multimodal research and highlights a fundamental misalignment between human-centric image quality and model-preferred input representations.

To investigate this, 
we systematically evaluate state-of-the-art MLLMs under 5 common degradation families (noise, motion blur, defocus blur, snow, fog) at multiple severities, spanning 13 vision-language datasets. 
Our findings reveal that not only do different models exhibit distinct quality preferences, but even within a single model, performance can vary across tasks or image instances in nontrivial ways. Moreover, we show that standard image restoration techniques, including strong pretrained pipelines and co-training approaches, are insufficient to recover or align with these idiosyncratic model behaviors.

To address this, we propose \textbf{Visual-Quality Test-Time Tuning (VQ-TTT)}, a lightweight and plug-and-play adaptation strategy that modulates the input image \textit{at test time}, without altering the backbone model or requiring additional training data. VQ-TTT consists of two components: (1) a learnable frequency-selective kernel layer inserted before the vision encoder, and (2) shallow-layer LoRA tuning for rapid adaptation. This design allows VQ-TTT to dynamically reshape image frequency content to better match the model’s task-specific preferences in a single forward pass.
Across extensive experiments, VQ-TTT demonstrates consistent and significant improvements over baseline and restoration-augmented MLLMs, achieving significant accuracy gain, while introducing negligible computational overhead. These results advocate for a paradigm shift: instead of relying on universally ``clean'' images, we argue for \textit{model-aligned visual adaptation} that tailors inputs to the unique preferences of each task and architecture.
Our main findings are summarized as follows:
 \begin{itemize}[leftmargin=*]
    \item \textbf{\textit{Empirical discovery of the visual-quality paradox}.}
Our analysis reveals that higher photographic fidelity does not uniformly benefit MLLMs; some tasks and models perform best on images that humans deem degraded, challenging prevailing assumptions.
    \item \textbf{\textit{Analysis behind the counterintuitive performance raise}.}
We employ relative attention and the logit lens technique to investigate how visual degradation influences model behavior, revealing that degradation can encourage MLLMs to focus more sharply and promote semantic alignment.
    \item \textbf{\textit{Restoration alone is insufficient}.}
Even strong pretrained restoration networks, when pipelined before an MLLM or co-trained with it, recover only a fraction of lost performance and occasionally hurt accuracy, underscoring the need for model-aligned adaptation.
    \item \textbf{\textit{VQ-TTT: task-adaptive visual-quality modulation}.}
We propose a plug-and-play test-time tuner—combining a learnable frequency-selective kernel with shallow-layer LoRA—that requires <1\% of original model parameters and no additional data.
    \item \textbf{\textit{State-of-the-art robustness gains with minimal cost}.}
VQ-TTT consistently boosts performance across all evaluated MLLMs, raising robustness scores by up to 8.6\% while adding negligible latency and memory overhead, and without altering downstream training pipelines.
\end{itemize}


\section{Preliminaries}

To investigate how the quality of visual input affects the performance of MLLMs, we conduct a comprehensive evaluation on VQA benchmarks using systematically corrupted image inputs. In this section, we first provide a brief overview of MLLMs, followed by a description of the relative attention and logit lens techniques employed for our analysis.

\paragraph{MLLM} MLLMs typically consist of three main components: a vision encoder, a projector, and an LLM backbone. Given a multimodal query $(x,v)$, where $x$ denotes the textual instruction and $v$ represents the input image, the vision encoder first processes the image $v$ into a sequence of image tokens. These tokens are then projected into the text embedding space by the projector and subsequently fed both the image and text tokens into the LLM backbone, which generates the final output in an autoregressive manner.

\paragraph{Relative Attention} Relative attention refers to the semantically normalized spatial attention from the answer to the image~\citep{zhang2025mllms}. Specifically, the spatial answer-to-image attention is denoted as the tensor product of the answer-to-token and token-to-image attention, $A_{si}^mk (x,v) = \hat{A}^m_{st} (x,v)\hat{A}^k_{ti}$(x,v) ($m,k$ is the layer indices of the LLM backbone and projector). Here, $\hat{A}^m_{st}(x,v)$ represents the average answer-to-token attention across all heads at LLM layer $m$, and $\hat{A}^k_{ti}(x,v)$ denotes the average token-to-image attention across all heads at projector layer $k$. To emphasize semantically relevant attention, the spatial attention is then normalized by a fixed reference instruction $x'=$``Write a general description of the image.'', i.e. $A_{rel} (x,v) = A_{si} (x,v) / A_{si} (x',v)$.

\paragraph{Logit Lens} Logit lens is a technique that decodes the hidden states from intermediate layers of a transformer into probability distributions over the vocabulary~\citep{logitlens}. This method enables interpretation of the model’s latent representations by revealing the information embedded at various depths of the network.

\section{Are VLMs Robust to Visual Quality?}


In this section, we evaluate the robustness and performance of MLLMs using Visual Question Answering (VQA) benchmarks under various visual degradations. We specifically examine three broad categories of image degradation: noise (Gaussian noise), blurring (motion blur and defocus blur), and adverse weather conditions (snow and fog). Each category introduces distinct distortion patterns reflecting real-world conditions: noise adds high-frequency artifacts, blurring removes high-frequency details, and weather conditions like snow and fog occlude critical visual information. By applying these diverse degradations, we aim to thoroughly assess the models' sensitivity and resilience to realistic image perturbations.

\subsection{Experimental Setup}

All image degradations are synthetically generated using the imagecorruptions library~\citep{michaelis2019dragon}. All experiments were performed on a computing cluster with 8× NVIDIA A6000 Ada GPUs. Detailed descriptions of the degradation methods, parameter settings, and implementation specifics are provided in the Appendix. 

\paragraph{Tasks and Datasets} 
To comprehensively evaluate the robustness and generalization ability of MLLMs under degraded visual conditions, we conduct experiments on several commonly adopted VQA benchmarks. Specifically, we utilize the following standardized VQA tasks and datasets:
\begin{itemize}
[leftmargin=1em,itemsep=0pt]
\vspace{-4pt}
    \item \textbf{MathVista~\citep{lu2024mathvista}:} a consolidated Mathematical reasoning benchmark with (1) seven mathematical reasoning types: algebraic reasoning (ALG), arithmetic reasoning (ARI), geometry reasoning (GEO), logical reasoning (LOG), numeric common sense (NUM), scientific reasoning (SCI), and statistical reasoning (STA); and (2) five primary tasks: figure question answering (FQA), geometry problem solving (GPS), math word problem (MWP), textbook question answering (TQA), and visual question answering (VQA). 

    \item \textbf{MMMU~\citep{yue2023mmmu}:} a comprehensive dataset comprises $\sim$11.5K multimodal questions sourced from authentic college exams, quizzes, and textbooks. It covers six broad academic disciplines—Art \& Design, Business, Science, Health \& Medicine, Humanities \& Social Science, and Technology \& Engineering—encompassing a total of 30 subjects and 183 detailed subfields.
    
    \item \textbf{ScienceQA~\citep{lu2022scienqa}:} a comprehensive benchmark that consists of $\sim$21K multimodal multiple-choice questions spanning diverse scientific domains—including natural science, social science, and language science. The questions are systematically organized into 26 high-level topics, 127 categories, and 379 distinct skills.
    
    \item \textbf{TextVQA~\citep{singh2019towards}:}  a novel dataset containing 8K multimodal open-ended, text-based questions that include the Optical Character Recognition (OCR) as a module, requiring the model to read and reason about the text in the image to be answered.

    \item \textbf{MME~\citep{Fu2023mme}:} a dataset that evaluates both perception (OCR, coarse-grained and fine-grained object recognition) and cognition (commonsense reasoning, numerical calculation, text translation, and code reasoning) abilities of MLLMs on a total of 14 subtasks.

\end{itemize}

\begin{table*}[htbp]
  \footnotesize
  \setlength{\tabcolsep}{5pt}
  \begin{center}
  \adjustbox{width=1.\linewidth}{
    \begin{tabular}{l|llllllll}
      \toprule
           \textbf{Model}  & MathVista & MMMU & ScienceQA$^T$ & ScienceQA$^I$  & TextVQA & MME$^P$ & MME$^C$       \\ 
      \midrule
      \rowcolor[gray]{0.95}LLaVA-v1.5-7B & 23.3  & 28.7 &  66.02 & 64.85 & 58.18 & 1510.28 & 357.85
      \\

      \rowcolor{gdmg!10}$+$Gaussian Noise & 24.2 & 28.6 & 66.38 & 65.59 & 56.53 & 1431.58 &  341.78
      \\
      $+$Gaussian Noise$+\mathcal{R_N}$ & 17.2 & 31.0 & 65.95 & 64.70 & 56.10 & 1419.18 & 345.00 \\  
      
      \rowcolor{gdmg!10}$+$Motion Blur & 24.4 & 29.5 & 66.12 & 65.05 & 54.33 & 1454.98 & 361.42 
      \\
        $+$Motion Blur$+\mathcal{R_N}$ & 17.8 & 31.0 & 66.14 & 65.10 & 56.77 & 1473.19 & 317.50 \\

      \rowcolor{gdmg!10}$+$Defocus Blur & 24.4 & 29.5 & 66.19 & 65.20 & 54.13 & 1435.93 & 326.07
      \\
      $+$Defocus Blur$+\mathcal{R_N}$ & 16.6 & 31.0 & 66.42 & 65.69 & 53.76 & 1435.69 & 347.86 \\

      \rowcolor{gdmg!10}$+$Snow & 24.1 & 28.2 & 65.36 & 63.46 & 53.24 & 1405.89 & 330.35 
      \\
      $+$Snow$+\mathcal{R_M}$ & 23.1 & 31.0 & 66.05 & 64.85 & 53.85 & 1391.24 & 315.36 & \\  

      \rowcolor{gdmg!10}$+$Fog & 24.8 & 28.6 & 65.39 & 63.51 & 57.13 & 1446.64 & 342.85
      \\
      $+$Fog$+\mathcal{R_M}$ & 23.8 & 31.0 & 65.83 & 64.40 & 56.68 & 1429.61 & 350.71 & \\  
      
      \midrule
      \rowcolor[gray]{0.95}LLaVa-v1.6-Mistral-7B & 26.5 & 34.2 & 76.02 & 71.34 & 63.80 & 1494.22 & 323.92
      \\

      \rowcolor{gdmg!10}$+$Gaussian Noise
      &29.8&33.7&76.44&72.24&59.09&1461.04&307.14\\

      $+$Gaussian Noise$+\mathcal{R_N}$&27.4&34.8&75.95&71.24&58.66&1438.57&303.57\\

      \rowcolor{gdmg!10}$+$Motion Blur&25.2&33.6&76.09&71.49&55.13&1454.99&339.64\\
      $+$Motion Blur$+\mathcal{R_N}$&27.8&22.3&76.09&71.54&59.35&1474.13&298.21\\

      \rowcolor{gdmg!10}$+$Defocus Blur
      &25.7& 34.2 &76.4&72.14&51.54&1395.71&275.71\\

      +Defocus Blur$+\mathcal{R_N}$&25.8&22.3&76.63&72.68&51.57&1371.05&285.36&
      \\


      \rowcolor{gdmg!10}$+$Snow & 28.7 & 33.7 & 76.00 & 71.29 & 55.39 & 1429.05 & 305.71
      \\
      $+$Snow$+\mathcal{R_M}$&26.5&22.3&76.11&71.59&56.38&1416.26&320
      \\
      \rowcolor{gdmg!10}$+$Fog & 30.7 & 34.2 & 76.00 & 71.29 & 62.65 & 1464.83 & 307.14
      \\
      $+$Fog$+\mathcal{R_M}$&25.4&34.8&76.23&71.84&61.16&1477.21&304.29
      \\
      \midrule
      \rowcolor[gray]{0.95}Qwen-2.5-VL-3B-instruct & 61.6 & 43.7 & 75.71 & 79.28 & 77.89 & 1515.32 & 615.00
      \\

      \rowcolor{gdmg!10}$+$Gaussian Noise & 57.5 & 42.1 & 73.90 & 75.81 & 65.50 & 1430.75 & 597.14
      \\
      $+$Gaussian Noise$+\mathcal{R_N}$ & 50.2 & 40.2 & 37.92 & 0.00 & 69.28 & 1439.65 & 580.00 \\  

      \rowcolor{gdmg!10}$+$Motion Blur & 56.6 & 41.4 & 74.02 & 75.81 & 61.39 & 1396.84 & 574.64
      \\
      $+$Motion Blur$+\mathcal{R_N}$ & 48.6 & 41.3 & 38.01 & 0.00 & 70.33 & 1502.61 & 557.50 \\  
 
      \rowcolor{gdmg!10}$+$Defocus Blur & 56.5 & 39.7 & 73.97 & 75.66 & 54.18 & 1346.74 & 555.35
      \\
      $+$Defocus Blur$+\mathcal{R_N}$ & 40.0 & 39.0 & 37.96 & 0.00 & 53.86 & 1399.06 & 530.71 \\  

      \rowcolor{gdmg!10}$+$Snow & 52.9 & 41.8 & 73.90 & 75.41 & 64.61 & 1427.12 & 554.28
      \\
      $+$Snow$+\mathcal{R_M}$ & 45.6 & 40.5 & 37.87 & 0.00 & 65.35 & 1443.55 & 581.43 \\  

      \rowcolor{gdmg!10}$+$Fog & 61.6 & 43.2 & 74.89 & 77.54 & 73.69 & 1499.79 & 605.71
      \\
      $+$Fog$+\mathcal{R_M}$ & 48.8 & 42.1 & 37.82 & 0.00 & 73.62 & 1478.08 & 626.43 \\  
      


 



      \bottomrule
    \end{tabular}}
  \end{center}
  \caption{Performance of MLLMs across common degradations and restorations (transformer-based). }
  \label{tab:degrad-eval-restore}
  \vspace{-5mm}
\end{table*}


\paragraph*{Pretrained Restoration} 
Intuitively, introducing degradations to the input images in a VQA benchmark is likely to cause a performance decline in MLLMs, as such perturbations compromise the quality of visual information and hinder the models' ability to accurately perceive and reason over the input. Furthermore, we further explore the effectiveness of integrating off-the-shelf pretrained image restoration models as a preprocessing step to recover corrupted visual inputs, investigating the extent to which these restoration models can mitigate the negative effects of image degradation and improve the downstream performance of MLLMs on VQA tasks. Specifically, we investigate two primary categories of restoration models—transformer-based and diffusion-based approaches—to restore degraded image inputs. The utilized models include:

\begin{itemize}[leftmargin=1em,itemsep=0pt]
\vspace{-4pt}
    \item \textbf{NAFNet~\citep{chen2022simple}} is a lightweight transformer-based image restoration model that replaces traiditonal nonlinear activations, such as ReLU, GELU, and softmax, with a SimpleGate mechanism for efficient image restoration. We adopt the official pretrained checkpoints for deblurring and denoising tasks, denoted as $\mathcal{R_N}$.
    \item \textbf{MWFormer~\citep{zhu2024mwformer}} introduces a weather-aware transformer framework tailored for real-world degradations such as snow and fog. It integrates degradation-aware mechanisms that adaptively handle different weather-induced distortions. In our experiment, we apply it to snow and fog, denoted as $\mathcal{R_M}$.
    \item \textbf{SUPIR~\citep{yu2024scaling}} is a diffusion-based model trained with 20M high-quality image-text pairs. It uses prompt-based guidance and learns to avoid visual artifacts by incorporating low-quality samples during training. We apply it for motion and defocus blur restoration, denoted as $\mathcal{R_S}$.
    \item \textbf{DiffBIR~\citep{lin2024diffbir}} introduces a two-staged restoration pipeline that decouples the restoration removal and information regeneration. It utilizes generative diffusion priors to reconstruct high-fidelity images. We use it for Gaussian noise restoration, denoted as $\mathcal{R_{DB}}$.
    \item \textbf{DA-CLIP~\citep{luo2023controlling}} integrates CLIP with a diffusion-based IR-SDE~\cite{luo2023image} framework using cross-attention for semantic guidance. It serves as an additional restoration model for snow and fog, denoted as $\mathcal{R_{DC}}$.
\end{itemize}
The results of the diffusion-based restoration models are included in the appendix.


\subsection{Results and Analysis}

In this section, we provide a comprehensive analysis of MLLM performance on VQA tasks with degraded image inputs. We also evaluate the effectiveness of applying existing image restoration models in improving model robustness under such conditions. We conduct experiments on VQA benchmarks with LLaVA-v1.5-7B~\citep{liu2024llava}, LLaVA-v1.6-Mistral-7B~\citep{llavanext}, and Qwen-2.5-VL-3B-instruct~\citep{bai2025qwen2}. The results are presented in the Table \ref{tab:degrad-eval-restore}. 

\paragraph*{The Paradox of High Image Quality} 
\begin{figure}[htbp]
    \centering
    \includegraphics[width=0.98\linewidth]{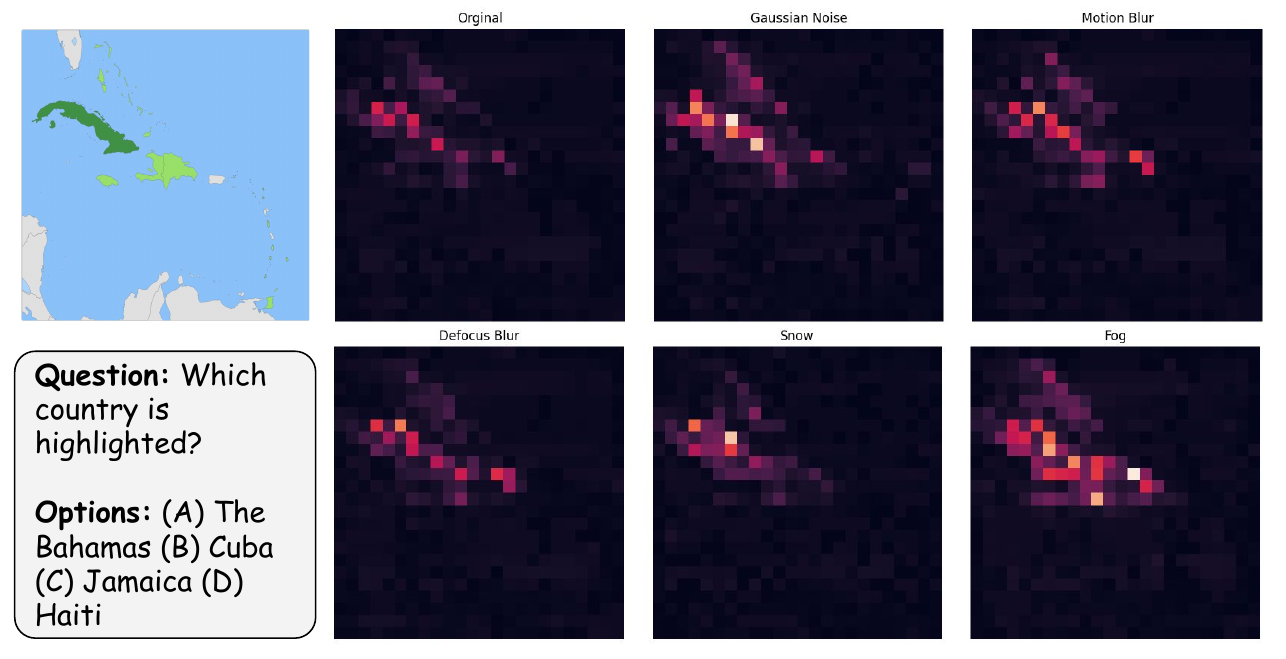}
    \caption{Visualizaion of the relative attention of LLaVA-v1.5-7B on a image-question pair (image 1301) of the ScienceQA dataset. Lighter regions indicate higher attention weights, while darker regions represent lower attention weights.}
    \label{fig:attn_map}   
\end{figure}

As shown in Table~\ref{tab:degrad-eval-restore}, existing MLLMs generally exhibit significant performance variability when presented with degraded image inputs, demonstrating their vulnerability to common visual perturbations. For tasks involving text and object recognition—such as TextVQA~\citep{singh2019textvqa} and the perception tasks in MME~\citep{Fu2023mme}—all evaluated models suffer notable performance drops. Among the various degradation types, MLLMs show relatively greater robustness to adverse weather corruptions, particularly fog, when performing recognition tasks. In contrast, blurring effects (i.e., motion blur and defocus blur) tend to cause the most severe degradation in performance, likely due to the loss of fine-grained spatial and textual details essential for accurate visual reasoning.

\textit{Surprisingly, MLLMs' performance does not always deteriorate when images deviate from human-perceived fidelity}. For cognitively demanding understanding and reasoning tasks (e.g., MathVista, ScienceQA), the introduction of certain degradations can, in some cases, lead to non-trivial performance improvements. For instance, LLaVA-v1.5-7B~\citep{llava} shows an average performance gain of 1.08 points on MathVista~\citep{lu2024mathvista} when evaluated on degraded images. Similarly, LLaVA-v1.6-Mistral-7B~\citep{llavanext} shows an average increase of 0.35 in image-based accuracy on ScienceQA~\citep{lu2022scienqa}. This counterintuitive phenomenon may suggest that certain types of visual perturbations may potentially guide the model’s attention toward salient features, suppressing irrelevant visual details and encouraging the model to focus on the essential semantic content required for reasoning. Alternatively, it may reflect a mismatch between human-centric fidelity metrics and the actual visual features exploited by MLLMs. To further investigate the underlying causes of this ``\textit{paradox}'', we conduct a detailed analysis of attention distributions, token-level predictions, and intermediate representations across different layers of the MLLMs under degraded image inputs.

\paragraph*{How Visual Perturbations Enhance MLLMs}

In this section, we employ the techniques of relative attention and logit lens to further analyze how image degradation affects the performance of MLLMs on cognitively demanding understanding and reasoning tasks. Our analysis aims to uncover underlying mechanisms that may explain the observed performance improvements under certain degraded conditions. 

\begin{wrapfigure}{r}{0.65\textwidth}
    \includegraphics[width=0.98\linewidth]{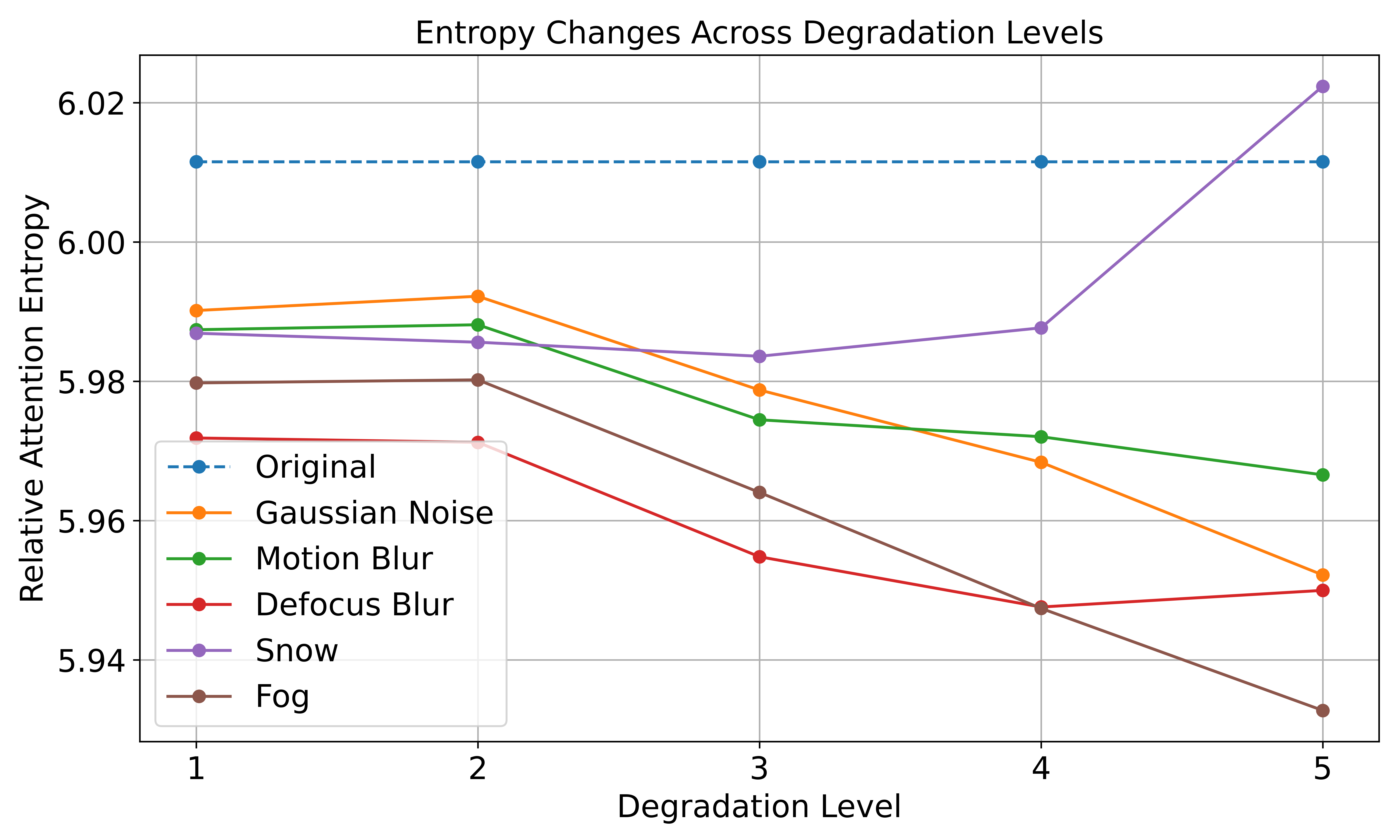}
    \caption{Changes in relative attention entropy of LLaVA-v1.5-7B on ScienceQA dataset across different levels of image degradation.}
    \label{fig:attn_entropy}   
\end{wrapfigure}
To investigate the underlying cause of the observed ``paradox'’, we use LLaVA-v1.5-7B~\citep{llava} as a representative example and visualize the relative attention heatmap, as illustrated in Figure~\ref{fig:attn_map}. Qualitatively, the heatmap illustrates the extent to which different image regions contribute to the generation of the final answer—effectively revealing where the MLLM ‘looks’ when answering the VQA. We can observe that the MLLM’s relative attention tends to \textit{concentrate more} on key regions relevant to answering the question. This suggests that the introduction of noise does not degrade the model’s ability to attend to important image areas; instead, it may even sharpen or intensify its focus. To further quantify this effect, we compute the entropy of the relative attention maps across five types of degradation, each evaluated at varying severity levels. As shown in Figure~\ref{fig:attn_entropy}, introducing image degradations in the ScienceQA dataset generally results in a decrease in relative attention entropy for LLaVA-v1.5-7B~\citep{llava}, supporting our qualitative case study findings, that MLLM tends to focus more on key image regions relevant to answering the question. Furthermore, as the severity level of degradation increases, the relative attention becomes even more concentrated, indicating a potential shift in the model’s strategy to cope with reduced visual fidelity by prioritizing semantically important regions.

\begin{figure}[htbp]
    \centering
    \includegraphics[width=0.275\linewidth]{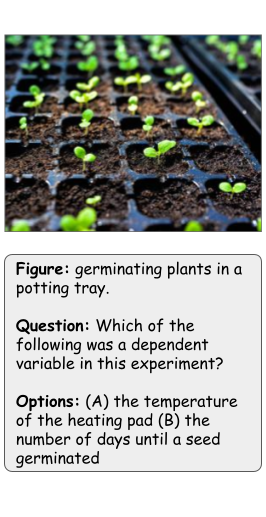}
    \includegraphics[width=0.33\linewidth]{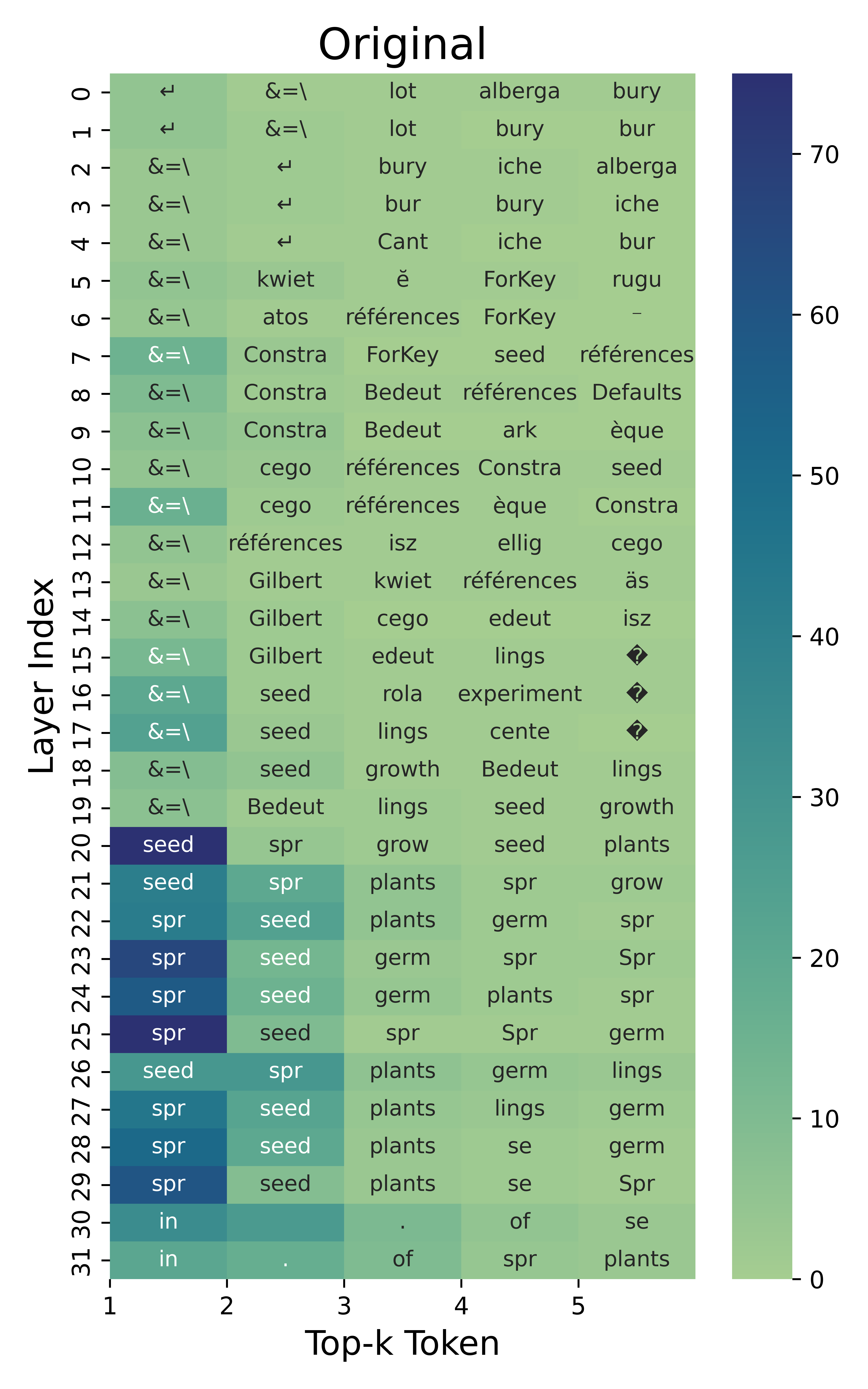}
    \includegraphics[width=0.33\linewidth]{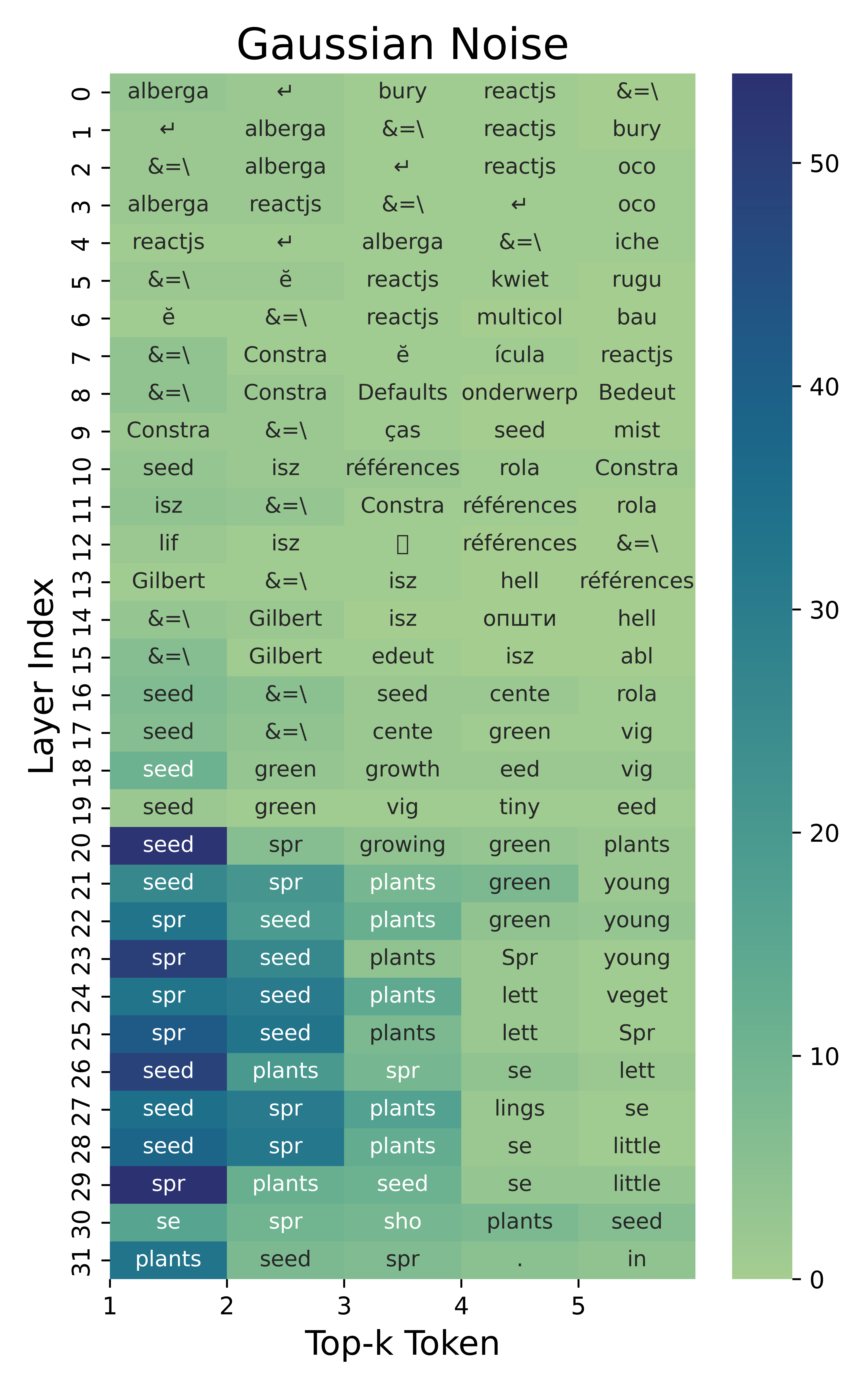}
    \caption{The logit lens of LLaVA-v1.5-7B model with a image-question pair (image 180) of the ScienceQA dataset. We present the next token distribution of the 50th image token for each layer in the heatmap. Lighter regions indicate higher probability, while darker regions represent lower probability.}
    \label{fig:logit-lens}   
\end{figure}


Furthermore, we apply the Logit Lens~\citep{logitlens} technique to analyze the behavior of MLLMs when presented with degraded image inputs. For consistency, we continue to use LLaVA-v1.5-7B~\citep{llava} as a representative case study, with additional results provided in the Appendix. Figure \ref{fig:logit-lens} presents the logit lens visualization for a germinating plant image-question pair, comparing the outputs when fed with the original image and a Gaussian-noised image. Semantically relevant tokens start emerging in shallow layers (Layers 7 to 9) for both input types, indicating that early-layer representations already encode meaningful semantic associations. Furthermore, as layer depth increases, the model exhibits greater confidence in predicting tokens that closely align with the image content, reflecting a progressive refinement of visual-semantic associations through deeper layers.

Notably, for the 50th image token, the most probable subsequent token is ``\textit{plants}''. Interestingly, the MLLM provided with the degraded image input successfully decodes this expected token, whereas the original image input results in decoding irrelevant tokens at the last layer. Although the logit lens method does not directly measure the visual understanding capability of the MLLM, these observations suggest that degraded images might unexpectedly guide the model toward stronger semantic coherence in certain contexts, highlighting intriguing dynamics in how MLLMs handle visual degradations.

\noindent\textbf{Degradation Then Restoration Aren’t a Cure-All}

In Table~\ref{tab:degrad-eval-restore} and the Appendix, we present the performance of MLLMs on VQA benchmarks using recovered image inputs, where the degraded images have been processed by off-the-shelf pretrained image restoration models. While these restoration models achieve favorable scores based on standard image quality assessment metrics, they do not consistently translate into improved performance for MLLMs. In some cases, for both perception and cognition tasks, the restored images even lead to worse performance than the directly degraded inputs (like ScienceQA and TextVQA for LLaVA-v1.5-7B).

This counter-intuitive finding indicates that current image restoration approaches optimize for perceptual quality metrics that do not align with the visual features most critical for MLLMs' reasoning capabilities. The restoration process appears to prioritize visual aesthetics over preserving the semantic content necessary for complex visual understanding tasks. While human perception may find the restored images visually improved, the MLLMs can struggle to extract meaningful features from them, indicating a mismatch between human visual quality assessment and machine vision preference.

\begin{tcolorbox}[before skip=2mm, after skip=0.0cm, boxsep=0.0cm, middle=0.0cm, top=0.1cm, bottom=0.1cm]
    \textit{\textbf{Takeaways:}
    \begin{itemize}[leftmargin=*]
        \item MLLMs’ performance does not always deteriorate when images deviate from human perceived fidelity, degradation can raise the MLLMs' performance on cognitively demanding understanding and reasoning tasks.
        \item Introducing visual degradation can help MLLMs focus more precisely on regions relevant to the input query, leading to improved semantic coherence.
        \item  Off-the-shelf pretrained image restoration models may not recover the visual features most important for MLLM, highlighting a mismatch between human-centric quality metrics and the visual cues leveraged by machine vision.
    \end{itemize}
    }
\end{tcolorbox}
\vspace*{0.2cm}

\section{How to Modulate Optimal Visual Quality for VLMs?}

To address the limitations of off-the-shelf image restoration approaches in enhancing MLLM performance on degraded inputs, we propose a simple yet effective test-time tuning method tailored for MLLMs in this section. 

\subsection{VQ-TTT: Visual-Quality Test-Time Tuning}

\noindent\textbf{Motivation.} Benchmark results reveal that, in certain scenarios, providing clearer visual inputs does not necessarily lead to improved performance in vision-language models (VLMs)—a phenomenon that is not coincidental. This observation suggests that the optimal visual input quality may vary depending on the specific downstream task or the architectural characteristics of the VLM. In other words, different tasks and models may exhibit distinct preferences for the frequency characteristics or levels of detail in the input images. To address this, we introduce a lightweight and cost-effective solution to dynamically realize such \textit{case-by-case preferences}. Specifically, we insert a learnable kernel before the frozen vision encoder, which enables adaptive modulation of the input image's visual quality in a task- or model-aware manner.

\begin{wrapfigure}{r}{0.6\textwidth}
    \includegraphics[width=0.98\linewidth]{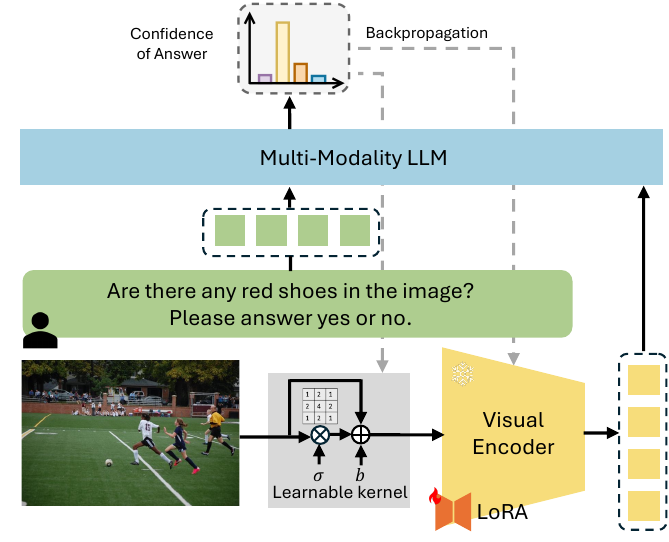}
    \caption{Illustration of VQ-TTT framework.}
    \label{fig:enter-label}
\end{wrapfigure}

\noindent\textbf{Learnable Kernel for Adaptive Visual Quality.} 
To modulate image quality, i.e., learnable blurring or sharpening, we introduce a \textbf{learnable kernel layer} that adaptively interpolates between the input image and its blurred version. Given an input image \( \mathbf{v} \), the modulated output is defined as:
\begin{equation}
\mathbf{v}' = (1 + b)\mathbf{v} - b \cdot (\mathbf{v} * \mathbf{K}_\sigma)
\label{eq:learnable_filter}
\end{equation}
where \( b \in \mathbb{R} \) is a learnable scalar blending coefficient, \( * \) denotes channel-wise (depthwise) convolution, and \( \mathbf{K}_\sigma \) is a separable Gaussian kernel with learnable standard deviation \( \sigma \). This simple yet flexible formulation enables the model to adaptively modulate the input image toward either sharpening (\( b > 0 \)) or blurring (\( b < 0 \)) using only two learnable parameters.


\noindent\textbf{Test-Time Visual Quality Modulation.}
To bridge the gap between the modulated image distribution and the original input distribution of vision-language models (VLMs), we introduce lightweight LoRA adapters into the shallow layers of the CLIP image encoder. Specifically, we insert LoRA modules into the first two layers, enabling adaptation with only a small number of parameters while keeping the original, well-pretrained weights of the VLM frozen.
To conduct a test-time tuning, given a VLM that predicts a distribution over textual outputs conditioned on an image-text pair, i.e., modulated image $\mathbf{v}'$ and text $\mathbf{t}$ pair, the entropy minimization loss can be defined as:
\begin{equation}
\mathcal{L}_{\text{entropy}} =  - \sum_{\mathbf{y} \in \mathcal{Y}} p_{\theta}(\mathbf{y} \mid \mathbf{v}', \mathbf{t}) \log p_{\theta}(\mathbf{y} \mid \mathbf{v}', \mathbf{t}) 
\end{equation}
where $p_{\theta}(\mathbf{y} \mid \mathbf{v}', \mathbf{t})$ is the model's predicted probability of output $\mathbf{y}$ given the input pair $(\mathbf{v}', \mathbf{t})$, with parameters ${\theta}$.
This objective encourages the model to produce confident, low-entropy predictions conditioned on multimodal inputs.

\subsection{Results and Analysis}
We adopt LLaVA series models as representative baselines to evaluate the effectiveness of our proposed VQ-TTT. Table~\ref{tab:ttt} presents a detailed comparison between the original baselines and their VQ-TTT-enhanced counterparts. Across a range of benchmark datasets, our method consistently achieves performance gains of up to 4.5\%, demonstrating its broad applicability. An exception is observed in the recognition metric on the MME dataset, where a minor decline occurs due to the inherent trade-off between perceptual quality and recognition accuracy. Notably, our test-time tuning strategy operates on an extremely lightweight module, comprising only two learnable parameters within a kernel layer and approximately 0.1M additional parameters from a plugged-in LoRA adapter. Full configuration details and parameter breakdowns are provided in the supplementary material. Importantly, VQ-TTT enables adaptive modulation of visual quality, allowing the model to dynamically tailor its outputs to the specific requirements of each task and evaluation dataset.

\begin{table*}[htbp]
  \footnotesize
  \setlength{\tabcolsep}{5pt}
  \begin{center}
  \adjustbox{width=1.\linewidth}{
    \begin{tabular}{l|llllllll}
      \toprule

      \textbf{Model}  & MathVista & MMMU & ScienceQA$^T$ & ScienceQA$^I$  & TextVQA & MME$^P$ & MME$^C$        \\ \hline
      \multicolumn{1}{l|}{LLaVA-v1.5-7B} 
      & 23.3 & 28.7 & 66.02 & 64.85& 58.18& 1500.13 & 316.43\\
       +VQ-TTT 
      & 24.4\textcolor{Green}{$_{\uparrow 1.1}$}
      & 	31.4\textcolor{Green}{$_{\uparrow 2.7}$}
      &   66.33\textcolor{Green}{$_{\uparrow 0.3}$}   & 65.44\textcolor{Green}{$_{\uparrow 0.6}$} & 	58.66\textcolor{Green}{$_{\uparrow 0.5}$}
      &	1512.23\textcolor{Green}{$_{\uparrow 12}$} & 307.86\textcolor{red}{$_{\downarrow 8.6}$}
      \\
        \midrule
            \multicolumn{1}{l|}{LLaVa-v1.6-Mistral-7B} 
      &26.5 & 34.2 & 76.02 & 71.34 & 63.80 & 1494.22 & 323.92\\

       +VQ-TTT 
      &27.5\textcolor{Green}{$_{\uparrow 1.0}$}
      & 	34.4\textcolor{Green}{$_{\uparrow 0.2}$}
      &	76.28\textcolor{Green}{$_{\uparrow 0.3}$}
      &	71.94\textcolor{Green}{$_{\uparrow 0.6}$}
      & 63.91\textcolor{Green}{$_{\uparrow 0.1}$}	 & 1503.88\textcolor{Green}{$_{\uparrow 9.6}$} & 318.93\textcolor{red}{$_{\downarrow 5.0}$}
      \\


      \bottomrule
    \end{tabular}}
  \end{center}
  \caption{Performance of VQ-TTT on LLaVA-v1.5-7B and LLaVa-v1.6-Mistral-7B.}
  \label{tab:ttt}
\end{table*}



\section{Related Work}
\subsection{Multimodal Large Language Models}



Multimodal Large Language Models (MLLMs)\citep{li2022blip, li2023blip2, liu2024llava, llavanext, llama3.2, Qwen-VL, Qwen2VL, lu2024deepseek, wu2024deepseek, bai2025qwen2, qwen2.5, beyer2024paligemma, abdin2024phi, abouelenin2025phi} build upon the language understanding and reasoning strengths of foundational LLMs\citep{devlin2018bert, radford2019gpt2, brown2020gpt3, team2023gemini, roziere2023codellama, touvron2023llama, touvron2023llama2, raffel2020t5, qwen2, qwen2.5} by incorporating visual processing capabilities. These models typically rely on visual encoders—such as CLIP~\citep{radford2021clip}—to transform image inputs into embedding representations that are then projected into the shared space of language embeddings. This cross-modal alignment has enabled a wide range of real-world applications, including biomedical image analysis~\citep{moor2023med, li2024llava-med}, autonomous driving~\citep{shao2024lmdrive, tian2024drivevlm, sima2023drivelm, openemma, wang2025generative, ma2025position, autotrust, luo2025v2x}, and embodied AI in robotics~\citep{rana2023sayplan, kim2024openvla, chen2025robo2vlm}.


\subsection{Test-Time Training in VLMs}
Test-time training (TTT) has emerged as a promising strategy to adapt pre-trained models to distribution shifts or downstream tasks without requiring full model retraining. While TTT has been extensively explored in vision~\cite{sun2020ttt, wang2021tent} and language~\cite{zhang2023tttllm} domains, its application to vision-language models (VLMs) remains relatively underexplored.
For VLMs, test-time adaptation is more challenging due to the dual-modal nature of inputs and the high computational cost of inference. Recent work has begun to address this: for instance,\cite{zhang2023vlmttt} proposes lightweight adapters for modulating vision-language alignment during inference, while\cite{li2024vpt} explores visual prompt tuning to adapt frozen VLMs to new visual domains. 

\subsection{Image Restoration and Co-Training}
Image restoration has been extensively studied to recover clean visual signals from degraded inputs, with applications spanning denoising, deblurring, and super-resolution~\cite{zhang2017beyond, dong2016image, zamir2021restormer}. Recent works have leveraged deep learning to achieve high-quality restoration, often optimizing perceptual fidelity using adversarial losses~\cite{ledig2017photo} or learned priors~\cite{ulyanov2018deep, saha2021image}. However, such methods typically aim to match human perception, assuming that cleaner images universally benefit downstream tasks.

Contrary to this assumption, recent studies suggest that restored images may not always align with task-specific model preferences, particularly in the context of vision-language models (VLMs)~\cite{bai2023align}. This observation motivates a shift from traditional restoration objectives toward task-aware or model-centric enhancement. Some recent approaches explore joint training of restoration and recognition networks~\cite{li2023joint, yang2020multi}, or co-training paradigms that integrate auxiliary tasks to improve robustness~\cite{zhang2021bytetrack}.

\section{Conclusion}
Our study reveals that higher perceptual visual quality does not always lead to better understanding in MLLMs, challenging conventional assumptions. Through a systematic analysis across diverse models and benchmarks, we uncover a visual-quality paradox, where degraded or stylistically altered inputs can unexpectedly enhance performance. To address this, we propose VQ-TTT, a lightweight and adaptive test-time tuning strategy that aligns input images with model-specific preferences using minimal learnable parameters. VQ-TTT delivers consistent gains across a wide range of models and tasks without relying on additional data or training. These results underscore the importance of rethinking input preparation in vision-language systems and advocate for adaptive, task-aware visual processing tailored to MLLMs’ internal biases.

\bibliography{main}

\begin{thebibliography}{80}
\providecommand{\natexlab}[1]{#1}
\providecommand{\url}[1]{\texttt{#1}}
\expandafter\ifx\csname urlstyle\endcsname\relax
  \providecommand{\doi}[1]{doi: #1}\else
  \providecommand{\doi}{doi: \begingroup \urlstyle{rm}\Url}\fi

\bibitem[Abdelhamed et~al.(2018)Abdelhamed, Lin, and Brown]{abdelhamed2018high}
A.~Abdelhamed, S.~Lin, and M.~S. Brown.
\newblock A high-quality denoising dataset for smartphone cameras.
\newblock In \emph{Proceedings of the IEEE conference on computer vision and pattern recognition}, pages 1692--1700, 2018.

\bibitem[Abdin et~al.(2024)Abdin, Aneja, Awadalla, Awadallah, Awan, Bach, Bahree, Bakhtiari, Bao, Behl, et~al.]{abdin2024phi}
M.~Abdin, J.~Aneja, H.~Awadalla, A.~Awadallah, A.~A. Awan, N.~Bach, A.~Bahree, A.~Bakhtiari, J.~Bao, H.~Behl, et~al.
\newblock Phi-3 technical report: A highly capable language model locally on your phone.
\newblock \emph{arXiv preprint arXiv:2404.14219}, 2024.

\bibitem[Abouelenin et~al.(2025)Abouelenin, Ashfaq, Atkinson, Awadalla, Bach, Bao, Benhaim, Cai, Chaudhary, Chen, et~al.]{abouelenin2025phi}
A.~Abouelenin, A.~Ashfaq, A.~Atkinson, H.~Awadalla, N.~Bach, J.~Bao, A.~Benhaim, M.~Cai, V.~Chaudhary, C.~Chen, et~al.
\newblock Phi-4-mini technical report: Compact yet powerful multimodal language models via mixture-of-loras.
\newblock \emph{arXiv preprint arXiv:2503.01743}, 2025.

\bibitem[Agrawal et~al.(2019)Agrawal, Desai, Wang, Chen, Jain, Johnson, Batra, Parikh, Lee, and Anderson]{agrawal2019nocaps}
H.~Agrawal, K.~Desai, Y.~Wang, X.~Chen, R.~Jain, M.~Johnson, D.~Batra, D.~Parikh, S.~Lee, and P.~Anderson.
\newblock Nocaps: Novel object captioning at scale.
\newblock In \emph{Proceedings of the IEEE/CVF international conference on computer vision}, pages 8948--8957, 2019.

\bibitem[Bai et~al.(2023{\natexlab{a}})Bai, Bai, Yang, Wang, Tan, Wang, Lin, Zhou, and Zhou]{Qwen-VL}
J.~Bai, S.~Bai, S.~Yang, S.~Wang, S.~Tan, P.~Wang, J.~Lin, C.~Zhou, and J.~Zhou.
\newblock Qwen-vl: A versatile vision-language model for understanding, localization, text reading, and beyond.
\newblock \emph{arXiv preprint arXiv:2308.12966}, 2023{\natexlab{a}}.

\bibitem[Bai et~al.(2025)Bai, Chen, Liu, Wang, Ge, Song, Dang, Wang, Wang, Tang, et~al.]{bai2025qwen2}
S.~Bai, K.~Chen, X.~Liu, J.~Wang, W.~Ge, S.~Song, K.~Dang, P.~Wang, S.~Wang, J.~Tang, et~al.
\newblock Qwen2. 5-vl technical report.
\newblock \emph{arXiv preprint arXiv:2502.13923}, 2025.

\bibitem[Bai et~al.(2023{\natexlab{b}})Bai, Zhang, Wang, Fu, Yuan, and Gao]{bai2023align}
Y.~Bai, X.~Zhang, Y.~Wang, Y.~Fu, L.~Yuan, and J.~Gao.
\newblock Align before generate: Vision-language pretraining with contrastive learning and knowledge distillation.
\newblock \emph{arXiv preprint arXiv:2302.14045}, 2023{\natexlab{b}}.

\bibitem[Beyer et~al.(2024)Beyer, Steiner, Pinto, Kolesnikov, Wang, Salz, Neumann, Alabdulmohsin, Tschannen, Bugliarello, et~al.]{beyer2024paligemma}
L.~Beyer, A.~Steiner, A.~S. Pinto, A.~Kolesnikov, X.~Wang, D.~Salz, M.~Neumann, I.~Alabdulmohsin, M.~Tschannen, E.~Bugliarello, et~al.
\newblock Paligemma: A versatile 3b vlm for transfer.
\newblock \emph{arXiv preprint arXiv:2407.07726}, 2024.

\bibitem[Brown et~al.(2020)Brown, Mann, Ryder, Subbiah, Kaplan, Dhariwal, Neelakantan, Shyam, Sastry, Askell, et~al.]{brown2020gpt3}
T.~Brown, B.~Mann, N.~Ryder, M.~Subbiah, J.~D. Kaplan, P.~Dhariwal, A.~Neelakantan, P.~Shyam, G.~Sastry, A.~Askell, et~al.
\newblock Language models are few-shot learners.
\newblock \emph{Advances in neural information processing systems}, 33:\penalty0 1877--1901, 2020.

\bibitem[Chen et~al.(2025)Chen, Xie, Ma, and Goldberg]{chen2025robo2vlm}
K.~Chen, S.~Xie, Z.~Ma, and K.~Goldberg.
\newblock Robo2vlm: Visual question answering from large-scale in-the-wild robot manipulation datasets.
\newblock \emph{arXiv preprint arXiv:2505.15517}, 2025.

\bibitem[Chen et~al.(2022)Chen, Chu, Zhang, and Sun]{chen2022simple}
L.~Chen, X.~Chu, X.~Zhang, and J.~Sun.
\newblock Simple baselines for image restoration.
\newblock In \emph{European conference on computer vision}, pages 17--33. Springer, 2022.

\bibitem[Chen et~al.(2015)Chen, Fang, Lin, Vedantam, Gupta, Doll{\'a}r, and Zitnick]{chen2015microsoftcoco}
X.~Chen, H.~Fang, T.-Y. Lin, R.~Vedantam, S.~Gupta, P.~Doll{\'a}r, and C.~L. Zitnick.
\newblock Microsoft coco captions: Data collection and evaluation server.
\newblock \emph{arXiv preprint arXiv:1504.00325}, 2015.

\bibitem[Devlin et~al.(2018)Devlin, Chang, Lee, and Toutanova]{devlin2018bert}
J.~Devlin, M.-W. Chang, K.~Lee, and K.~Toutanova.
\newblock Bert: Pre-training of deep bidirectional transformers for language understanding.
\newblock \emph{arXiv preprint arXiv:1810.04805}, 2018.

\bibitem[Dong et~al.(2016)Dong, Loy, He, and Tang]{dong2016image}
C.~Dong, C.~C. Loy, K.~He, and X.~Tang.
\newblock Image super-resolution using deep convolutional networks.
\newblock In \emph{IEEE Transactions on Pattern Analysis and Machine Intelligence}, volume~38, pages 295--307. IEEE, 2016.

\bibitem[Fu et~al.(2023)Fu, Chen, Shen, Qin, Zhang, Lin, Yang, Zheng, Li, Sun, Wu, and Ji]{Fu2023mme}
C.~Fu, P.~Chen, Y.~Shen, Y.~Qin, M.~Zhang, X.~Lin, J.~Yang, X.~Zheng, K.~Li, X.~Sun, Y.~Wu, and R.~Ji.
\newblock {MME: A Comprehensive Evaluation Benchmark for Multimodal Large Language Models}.
\newblock \emph{arXiv}, June 2023.
\newblock \doi{10.48550/arXiv.2306.13394}.

\bibitem[Gurari et~al.(2018)Gurari, Li, Stangl, Guo, Lin, Grauman, Luo, and Bigham]{gurari2018vizwiz}
D.~Gurari, Q.~Li, A.~J. Stangl, A.~Guo, C.~Lin, K.~Grauman, J.~Luo, and J.~P. Bigham.
\newblock Vizwiz grand challenge: Answering visual questions from blind people.
\newblock In \emph{Proceedings of the IEEE conference on computer vision and pattern recognition}, pages 3608--3617, 2018.

\bibitem[Hu et~al.(2025)Hu, Wang, Xing, Chen, and Tu]{hu2025mrag}
C.-W. Hu, Y.~Wang, S.~Xing, C.-J. Chen, and Z.~Tu.
\newblock mrag: Elucidating the design space of multi-modal retrieval-augmented generation.
\newblock \emph{arXiv preprint arXiv:2505.24073}, 2025.

\bibitem[Hudson and Manning(2019)]{hudson2019gqa}
D.~A. Hudson and C.~D. Manning.
\newblock Gqa: A new dataset for real-world visual reasoning and compositional question answering.
\newblock In \emph{Proceedings of the IEEE/CVF conference on computer vision and pattern recognition}, pages 6700--6709, 2019.

\bibitem[Kim et~al.(2024)Kim, Pertsch, Karamcheti, Xiao, Balakrishna, Nair, Rafailov, Foster, Lam, Sanketi, et~al.]{kim2024openvla}
M.~J. Kim, K.~Pertsch, S.~Karamcheti, T.~Xiao, A.~Balakrishna, S.~Nair, R.~Rafailov, E.~Foster, G.~Lam, P.~Sanketi, et~al.
\newblock Openvla: An open-source vision-language-action model.
\newblock \emph{arXiv preprint arXiv:2406.09246}, 2024.

\bibitem[Ledig et~al.(2017)Ledig, Theis, Husz{\'a}r, Caballero, Cunningham, Acosta, Aitken, Tejani, Totz, Wang, and Shi]{ledig2017photo}
C.~Ledig, L.~Theis, F.~Husz{\'a}r, J.~Caballero, A.~Cunningham, A.~Acosta, A.~Aitken, A.~Tejani, J.~Totz, Z.~Wang, and W.~Shi.
\newblock Photo-realistic single image super-resolution using a generative adversarial network.
\newblock \emph{CVPR}, 2017.

\bibitem[Li et~al.(2024{\natexlab{a}})Li, Wong, Zhang, Usuyama, Liu, Yang, Naumann, Poon, and Gao]{li2024llava-med}
C.~Li, C.~Wong, S.~Zhang, N.~Usuyama, H.~Liu, J.~Yang, T.~Naumann, H.~Poon, and J.~Gao.
\newblock Llava-med: Training a large language-and-vision assistant for biomedicine in one day.
\newblock \emph{Advances in Neural Information Processing Systems}, 36, 2024{\natexlab{a}}.

\bibitem[Li et~al.(2023{\natexlab{a}})Li, Wang, Sun, Van~Gool, and Timofte]{li2023joint}
F.~Li, Y.~Wang, Q.~Sun, L.~Van~Gool, and R.~Timofte.
\newblock Joint image restoration and recognition using a single model.
\newblock In \emph{CVPR}, 2023{\natexlab{a}}.

\bibitem[Li et~al.(2024{\natexlab{b}})Li, Zhang, Zhang, Zhang, Li, Li, Ma, and Li]{llavanext}
F.~Li, R.~Zhang, H.~Zhang, Y.~Zhang, B.~Li, W.~Li, Z.~Ma, and C.~Li.
\newblock Llava-next-interleave: Tackling multi-image, video, and 3d in large multimodal models.
\newblock \emph{arXiv preprint arXiv:2407.07895}, 2024{\natexlab{b}}.

\bibitem[Li et~al.(2024{\natexlab{c}})Li, Zhang, Liu, Wang, Wu, and Wang]{li2024vpt}
H.~Li, Z.~Zhang, J.~Liu, X.~Wang, J.~Wu, and Y.~Wang.
\newblock Visual prompt tuning for adapting vision-language models.
\newblock In \emph{Proceedings of the IEEE/CVF Conference on Computer Vision and Pattern Recognition (CVPR)}, 2024{\natexlab{c}}.

\bibitem[Li et~al.(2022)Li, Li, Xiong, and Hoi]{li2022blip}
J.~Li, D.~Li, C.~Xiong, and S.~Hoi.
\newblock Blip: Bootstrapping language-image pre-training for unified vision-language understanding and generation.
\newblock In \emph{International conference on machine learning}, pages 12888--12900. PMLR, 2022.

\bibitem[Li et~al.(2023{\natexlab{b}})Li, Li, Savarese, and Hoi]{li2023blip2}
J.~Li, D.~Li, S.~Savarese, and S.~Hoi.
\newblock Blip-2: Bootstrapping language-image pre-training with frozen image encoders and large language models.
\newblock In \emph{International conference on machine learning}, pages 19730--19742. PMLR, 2023{\natexlab{b}}.

\bibitem[Li et~al.(2025)Li, Zou, Wu, Li, Xing, Zheng, Hu, Wang, Li, Yuan, et~al.]{li2025safeflow}
P.~Li, X.~Zou, Z.~Wu, R.~Li, S.~Xing, H.~Zheng, Z.~Hu, Y.~Wang, H.~Li, Q.~Yuan, et~al.
\newblock Safeflow: A principled protocol for trustworthy and transactional autonomous agent systems.
\newblock \emph{arXiv preprint arXiv:2506.07564}, 2025.

\bibitem[Lin et~al.(2024)Lin, He, Chen, Lyu, Dai, Yu, Qiao, Ouyang, and Dong]{lin2024diffbir}
X.~Lin, J.~He, Z.~Chen, Z.~Lyu, B.~Dai, F.~Yu, Y.~Qiao, W.~Ouyang, and C.~Dong.
\newblock Diffbir: Toward blind image restoration with generative diffusion prior.
\newblock In \emph{European Conference on Computer Vision}, pages 430--448. Springer, 2024.

\bibitem[Liu et~al.(2023)Liu, Li, Wu, and Lee]{llava}
H.~Liu, C.~Li, Q.~Wu, and Y.~J. Lee.
\newblock Visual instruction tuning.
\newblock \emph{arXiv preprint arXiv:2304.08485}, 2023.
\newblock URL \url{https://arxiv.org/abs/2304.08485}.

\bibitem[Liu et~al.(2024)Liu, Li, Wu, and Lee]{liu2024llava}
H.~Liu, C.~Li, Q.~Wu, and Y.~J. Lee.
\newblock Visual instruction tuning.
\newblock \emph{Advances in neural information processing systems}, 36, 2024.

\bibitem[Lu et~al.(2024{\natexlab{a}})Lu, Liu, Zhang, Wang, Dong, Liu, Sun, Ren, Li, Yang, et~al.]{lu2024deepseek}
H.~Lu, W.~Liu, B.~Zhang, B.~Wang, K.~Dong, B.~Liu, J.~Sun, T.~Ren, Z.~Li, H.~Yang, et~al.
\newblock Deepseek-vl: towards real-world vision-language understanding.
\newblock \emph{arXiv preprint arXiv:2403.05525}, 2024{\natexlab{a}}.

\bibitem[Lu et~al.(2022)Lu, Mishra, Xia, Qiu, Chang, Zhu, Tafjord, Clark, and Kalyan]{lu2022scienqa}
P.~Lu, S.~Mishra, T.~Xia, L.~Qiu, K.-W. Chang, S.-C. Zhu, O.~Tafjord, P.~Clark, and A.~Kalyan.
\newblock Learn to explain: Multimodal reasoning via thought chains for science question answering.
\newblock In \emph{The 36th Conference on Neural Information Processing Systems (NeurIPS)}, 2022.

\bibitem[Lu et~al.(2024{\natexlab{b}})Lu, Bansal, Xia, Liu, Li, Hajishirzi, Cheng, Chang, Galley, and Gao]{lu2024mathvista}
P.~Lu, H.~Bansal, T.~Xia, J.~Liu, C.~Li, H.~Hajishirzi, H.~Cheng, K.-W. Chang, M.~Galley, and J.~Gao.
\newblock Mathvista: Evaluating mathematical reasoning of foundation models in visual contexts.
\newblock In \emph{International Conference on Learning Representations (ICLR)}, 2024{\natexlab{b}}.

\bibitem[Luo et~al.(2025)Luo, Yang, Ding, Gao, Xing, Zhou, Tu, and Liu]{luo2025v2x}
X.~Luo, F.~Yang, F.~Ding, X.~Gao, S.~Xing, Y.~Zhou, Z.~Tu, and C.~Liu.
\newblock V2x-unipool: Unifying multimodal perception and knowledge reasoning for autonomous driving.
\newblock \emph{arXiv preprint arXiv:2506.02580}, 2025.

\bibitem[Luo et~al.(2023{\natexlab{a}})Luo, Gustafsson, Zhao, Sj{\"o}lund, and Sch{\"o}n]{luo2023controlling}
Z.~Luo, F.~K. Gustafsson, Z.~Zhao, J.~Sj{\"o}lund, and T.~B. Sch{\"o}n.
\newblock Controlling vision-language models for multi-task image restoration.
\newblock \emph{arXiv preprint arXiv:2310.01018}, 2023{\natexlab{a}}.

\bibitem[Luo et~al.(2023{\natexlab{b}})Luo, Gustafsson, Zhao, Sj{\"o}lund, and Sch{\"o}n]{luo2023image}
Z.~Luo, F.~K. Gustafsson, Z.~Zhao, J.~Sj{\"o}lund, and T.~B. Sch{\"o}n.
\newblock Image restoration with mean-reverting stochastic differential equations.
\newblock \emph{arXiv preprint arXiv:2301.11699}, 2023{\natexlab{b}}.

\bibitem[Ma et~al.(2025)Ma, Ye, Cui, Zhang, Xing, Ke, Wang, Miao, Chen, Rezatofighi, et~al.]{ma2025position}
Y.~Ma, W.~Ye, C.~Cui, H.~Zhang, S.~Xing, F.~Ke, J.~Wang, C.~Miao, J.~Chen, H.~Rezatofighi, et~al.
\newblock Position: Prospective of autonomous driving-multimodal llms world models embodied intelligence ai alignment and mamba.
\newblock In \emph{Proceedings of the Winter Conference on Applications of Computer Vision}, pages 1010--1026, 2025.

\bibitem[Meta(2024)]{llama3.2}
Meta.
\newblock Llama 3.2: Revolutionizing edge ai and vision with open, customizable models.
\newblock 2024.
\newblock URL \url{https://ai.meta.com/blog/llama-3-2-connect-2024-vision-edge-mobile-devices/}.

\bibitem[Michaelis et~al.(2019)Michaelis, Mitzkus, Geirhos, Rusak, Bringmann, Ecker, Bethge, and Brendel]{michaelis2019dragon}
C.~Michaelis, B.~Mitzkus, R.~Geirhos, E.~Rusak, O.~Bringmann, A.~S. Ecker, M.~Bethge, and W.~Brendel.
\newblock Benchmarking robustness in object detection: Autonomous driving when winter is coming.
\newblock \emph{arXiv preprint arXiv:1907.07484}, 2019.

\bibitem[Moor et~al.(2023)Moor, Huang, Wu, Yasunaga, Dalmia, Leskovec, Zakka, Reis, and Rajpurkar]{moor2023med}
M.~Moor, Q.~Huang, S.~Wu, M.~Yasunaga, Y.~Dalmia, J.~Leskovec, C.~Zakka, E.~P. Reis, and P.~Rajpurkar.
\newblock Med-flamingo: a multimodal medical few-shot learner.
\newblock In \emph{Machine Learning for Health (ML4H)}, pages 353--367. PMLR, 2023.

\bibitem[Nah et~al.(2017)Nah, Hyun~Kim, and Mu~Lee]{nah2017deep}
S.~Nah, T.~Hyun~Kim, and K.~Mu~Lee.
\newblock Deep multi-scale convolutional neural network for dynamic scene deblurring.
\newblock In \emph{Proceedings of the IEEE conference on computer vision and pattern recognition}, pages 3883--3891, 2017.

\bibitem[nostalgebraist(2020)]{logitlens}
nostalgebraist.
\newblock interpreting gpt: the logit lens.
\newblock 2020.
\newblock URL \url{https://www.lesswrong.com/posts/AcKRB8wDpdaN6v6ru/interpreting-gpt-the-logit-lens}.

\bibitem[Radford et~al.(2019)Radford, Wu, Child, Luan, Amodei, Sutskever, et~al.]{radford2019gpt2}
A.~Radford, J.~Wu, R.~Child, D.~Luan, D.~Amodei, I.~Sutskever, et~al.
\newblock Language models are unsupervised multitask learners.
\newblock \emph{OpenAI blog}, 1\penalty0 (8):\penalty0 9, 2019.

\bibitem[Radford et~al.(2021)Radford, Kim, Hallacy, Ramesh, Goh, Agarwal, Sastry, Askell, Mishkin, Clark, et~al.]{radford2021clip}
A.~Radford, J.~W. Kim, C.~Hallacy, A.~Ramesh, G.~Goh, S.~Agarwal, G.~Sastry, A.~Askell, P.~Mishkin, J.~Clark, et~al.
\newblock Learning transferable visual models from natural language supervision.
\newblock In \emph{International conference on machine learning}, pages 8748--8763. PMLR, 2021.

\bibitem[Raffel et~al.(2020)Raffel, Shazeer, Roberts, Lee, Narang, Matena, Zhou, Li, and Liu]{raffel2020t5}
C.~Raffel, N.~Shazeer, A.~Roberts, K.~Lee, S.~Narang, M.~Matena, Y.~Zhou, W.~Li, and P.~J. Liu.
\newblock Exploring the limits of transfer learning with a unified text-to-text transformer.
\newblock \emph{Journal of machine learning research}, 21\penalty0 (140):\penalty0 1--67, 2020.

\bibitem[Rana et~al.(2023)Rana, Haviland, Garg, Abou-Chakra, Reid, and Suenderhauf]{rana2023sayplan}
K.~Rana, J.~Haviland, S.~Garg, J.~Abou-Chakra, I.~Reid, and N.~Suenderhauf.
\newblock Sayplan: Grounding large language models using 3d scene graphs for scalable robot task planning.
\newblock In \emph{7th Annual Conference on Robot Learning}, 2023.

\bibitem[Roziere et~al.(2023)Roziere, Gehring, Gloeckle, Sootla, Gat, Tan, Adi, Liu, Remez, Rapin, et~al.]{roziere2023codellama}
B.~Roziere, J.~Gehring, F.~Gloeckle, S.~Sootla, I.~Gat, X.~E. Tan, Y.~Adi, J.~Liu, T.~Remez, J.~Rapin, et~al.
\newblock Code llama: Open foundation models for code.
\newblock \emph{arXiv preprint arXiv:2308.12950}, 2023.

\bibitem[Saha et~al.(2021)Saha, Sinha, and Bandyopadhyay]{saha2021image}
S.~Saha, A.~Sinha, and S.~Bandyopadhyay.
\newblock Image restoration using very deep convolutional encoder-decoder networks with symmetric skip connections.
\newblock In \emph{Pattern Recognition Letters}, volume 138, pages 185--191. Elsevier, 2021.

\bibitem[Shao et~al.(2024)Shao, Hu, Wang, Song, Waslander, Liu, and Li]{shao2024lmdrive}
H.~Shao, Y.~Hu, L.~Wang, G.~Song, S.~L. Waslander, Y.~Liu, and H.~Li.
\newblock Lmdrive: Closed-loop end-to-end driving with large language models.
\newblock In \emph{Proceedings of the IEEE/CVF Conference on Computer Vision and Pattern Recognition}, pages 15120--15130, 2024.

\bibitem[Sima et~al.(2023)Sima, Renz, Chitta, Chen, Zhang, Xie, Luo, Geiger, and Li]{sima2023drivelm}
C.~Sima, K.~Renz, K.~Chitta, L.~Chen, H.~Zhang, C.~Xie, P.~Luo, A.~Geiger, and H.~Li.
\newblock Drivelm: Driving with graph visual question answering.
\newblock \emph{arXiv preprint arXiv:2312.14150}, 2023.

\bibitem[Singh et~al.(2019{\natexlab{a}})Singh, Natarajan, Shah, Jiang, Chen, Batra, Parikh, and Rohrbach]{singh2019textvqa}
A.~Singh, V.~Natarajan, M.~Shah, Y.~Jiang, X.~Chen, D.~Batra, D.~Parikh, and M.~Rohrbach.
\newblock Towards vqa models that can read.
\newblock In \emph{Proceedings of the IEEE/CVF conference on computer vision and pattern recognition}, pages 8317--8326, 2019{\natexlab{a}}.

\bibitem[Singh et~al.(2019{\natexlab{b}})Singh, Natarajan, Shah, Jiang, Chen, Batra, Parikh, and Rohrbach]{singh2019towards}
A.~Singh, V.~Natarajan, M.~Shah, Y.~Jiang, X.~Chen, D.~Batra, D.~Parikh, and M.~Rohrbach.
\newblock Towards vqa models that can read.
\newblock In \emph{Proceedings of the IEEE/CVF conference on computer vision and pattern recognition}, pages 8317--8326, 2019{\natexlab{b}}.

\bibitem[Sun et~al.(2020)Sun, Tzamarias, and Schiele]{sun2020ttt}
Q.~Sun, A.~Tzamarias, and B.~Schiele.
\newblock Test-time training with self-supervision for generalization under distribution shifts.
\newblock In \emph{International Conference on Machine Learning (ICML)}, pages 9229--9248, 2020.

\bibitem[Team et~al.(2023)Team, Anil, Borgeaud, Wu, Alayrac, Yu, Soricut, Schalkwyk, Dai, Hauth, et~al.]{team2023gemini}
G.~Team, R.~Anil, S.~Borgeaud, Y.~Wu, J.-B. Alayrac, J.~Yu, R.~Soricut, J.~Schalkwyk, A.~M. Dai, A.~Hauth, et~al.
\newblock Gemini: a family of highly capable multimodal models.
\newblock \emph{arXiv preprint arXiv:2312.11805}, 2023.

\bibitem[Team(2024)]{qwen2.5}
Q.~Team.
\newblock Qwen2.5: A party of foundation models, September 2024.
\newblock URL \url{https://qwenlm.github.io/blog/qwen2.5/}.

\bibitem[Tian et~al.(2024)Tian, Gu, Li, Liu, Hu, Wang, Zhan, Jia, Lang, and Zhao]{tian2024drivevlm}
X.~Tian, J.~Gu, B.~Li, Y.~Liu, C.~Hu, Y.~Wang, K.~Zhan, P.~Jia, X.~Lang, and H.~Zhao.
\newblock Drivevlm: The convergence of autonomous driving and large vision-language models.
\newblock \emph{arXiv preprint arXiv:2402.12289}, 2024.

\bibitem[Touvron et~al.(2023{\natexlab{a}})Touvron, Lavril, Izacard, Martinet, Lachaux, Lacroix, Rozi{\`e}re, Goyal, Hambro, Azhar, et~al.]{touvron2023llama}
H.~Touvron, T.~Lavril, G.~Izacard, X.~Martinet, M.-A. Lachaux, T.~Lacroix, B.~Rozi{\`e}re, N.~Goyal, E.~Hambro, F.~Azhar, et~al.
\newblock Llama: Open and efficient foundation language models.
\newblock \emph{arXiv preprint arXiv:2302.13971}, 2023{\natexlab{a}}.

\bibitem[Touvron et~al.(2023{\natexlab{b}})Touvron, Martin, Stone, Albert, Almahairi, Babaei, Bashlykov, Batra, Bhargava, Bhosale, et~al.]{touvron2023llama2}
H.~Touvron, L.~Martin, K.~Stone, P.~Albert, A.~Almahairi, Y.~Babaei, N.~Bashlykov, S.~Batra, P.~Bhargava, S.~Bhosale, et~al.
\newblock Llama 2: Open foundation and fine-tuned chat models.
\newblock \emph{arXiv preprint arXiv:2307.09288}, 2023{\natexlab{b}}.

\bibitem[Ulyanov et~al.(2018)Ulyanov, Vedaldi, and Lempitsky]{ulyanov2018deep}
D.~Ulyanov, A.~Vedaldi, and V.~Lempitsky.
\newblock Deep image prior.
\newblock \emph{CVPR}, 2018.

\bibitem[Wang et~al.(2021)Wang, Bao, Dong, Zhu, and Gonzalez]{wang2021tent}
D.~Wang, J.~Bao, X.~Dong, J.-Y. Zhu, and J.~E. Gonzalez.
\newblock Tent: Fully test-time adaptation by entropy minimization.
\newblock In \emph{International Conference on Learning Representations (ICLR)}, 2021.

\bibitem[Wang et~al.(2024)Wang, Bai, Tan, Wang, Fan, Bai, Chen, Liu, Wang, Ge, Fan, Dang, Du, Ren, Men, Liu, Zhou, Zhou, and Lin]{Qwen2VL}
P.~Wang, S.~Bai, S.~Tan, S.~Wang, Z.~Fan, J.~Bai, K.~Chen, X.~Liu, J.~Wang, W.~Ge, Y.~Fan, K.~Dang, M.~Du, X.~Ren, R.~Men, D.~Liu, C.~Zhou, J.~Zhou, and J.~Lin.
\newblock Qwen2-vl: Enhancing vision-language model's perception of the world at any resolution.
\newblock \emph{arXiv preprint arXiv:2409.12191}, 2024.

\bibitem[Wang et~al.(2025)Wang, Xing, Can, Li, Hua, Tian, Mo, Gao, Wu, Zhou, et~al.]{wang2025generative}
Y.~Wang, S.~Xing, C.~Can, R.~Li, H.~Hua, K.~Tian, Z.~Mo, X.~Gao, K.~Wu, S.~Zhou, et~al.
\newblock Generative ai for autonomous driving: Frontiers and opportunities.
\newblock \emph{arXiv preprint arXiv:2505.08854}, 2025.

\bibitem[Wu et~al.(2024)Wu, Chen, Pan, Liu, Liu, Dai, Gao, Ma, Wu, Wang, et~al.]{wu2024deepseek}
Z.~Wu, X.~Chen, Z.~Pan, X.~Liu, W.~Liu, D.~Dai, H.~Gao, Y.~Ma, C.~Wu, B.~Wang, et~al.
\newblock Deepseek-vl2: Mixture-of-experts vision-language models for advanced multimodal understanding.
\newblock \emph{arXiv preprint arXiv:2412.10302}, 2024.

\bibitem[Xing et~al.(2024{\natexlab{a}})Xing, Hua, Gao, Zhu, Li, Tian, Li, Huang, Yang, Wang, Zhou, Yao, and Tu]{autotrust}
S.~Xing, H.~Hua, X.~Gao, S.~Zhu, R.~Li, K.~Tian, X.~Li, H.~Huang, T.~Yang, Z.~Wang, Y.~Zhou, H.~Yao, and Z.~Tu.
\newblock {AutoTrust: Benchmarking Trustworthiness in Large Vision Language Models for Autonomous Driving}.
\newblock \emph{arXiv}, Dec. 2024{\natexlab{a}}.
\newblock \doi{10.48550/arXiv.2412.15206}.

\bibitem[Xing et~al.(2024{\natexlab{b}})Xing, Qian, Wang, Hua, Tian, Zhou, and Tu]{openemma}
S.~Xing, C.~Qian, Y.~Wang, H.~Hua, K.~Tian, Y.~Zhou, and Z.~Tu.
\newblock Openemma: Open-source multimodal model for end-to-end autonomous driving.
\newblock \emph{arXiv}, Dec. 2024{\natexlab{b}}.
\newblock \doi{10.48550/arXiv.2412.15208}.

\bibitem[Xing et~al.(2025{\natexlab{a}})Xing, Sun, Xie, Chen, Huang, Wang, Li, Song, and Tu]{xing2025can}
S.~Xing, Z.~Sun, S.~Xie, K.~Chen, Y.~Huang, Y.~Wang, J.~Li, D.~Song, and Z.~Tu.
\newblock Can large vision language models read maps like a human?
\newblock \emph{arXiv preprint arXiv:2503.14607}, 2025{\natexlab{a}}.

\bibitem[Xing et~al.(2025{\natexlab{b}})Xing, Wang, Li, Bai, Wang, Qian, Yao, and Tu]{xing2025re}
S.~Xing, Y.~Wang, P.~Li, R.~Bai, Y.~Wang, C.~Qian, H.~Yao, and Z.~Tu.
\newblock Re-align: Aligning vision language models via retrieval-augmented direct preference optimization.
\newblock \emph{arXiv preprint arXiv:2502.13146}, 2025{\natexlab{b}}.

\bibitem[Yang et~al.(2024)Yang, Yang, Hui, Zheng, Yu, Zhou, Li, Li, Liu, Huang, Dong, Wei, Lin, Tang, Wang, Yang, Tu, Zhang, Ma, Xu, Zhou, Bai, He, Lin, Dang, Lu, Chen, Yang, Li, Xue, Ni, Zhang, Wang, Peng, Men, Gao, Lin, Wang, Bai, Tan, Zhu, Li, Liu, Ge, Deng, Zhou, Ren, Zhang, Wei, Ren, Fan, Yao, Zhang, Wan, Chu, Liu, Cui, Zhang, and Fan]{qwen2}
A.~Yang, B.~Yang, B.~Hui, B.~Zheng, B.~Yu, C.~Zhou, C.~Li, C.~Li, D.~Liu, F.~Huang, G.~Dong, H.~Wei, H.~Lin, J.~Tang, J.~Wang, J.~Yang, J.~Tu, J.~Zhang, J.~Ma, J.~Xu, J.~Zhou, J.~Bai, J.~He, J.~Lin, K.~Dang, K.~Lu, K.~Chen, K.~Yang, M.~Li, M.~Xue, N.~Ni, P.~Zhang, P.~Wang, R.~Peng, R.~Men, R.~Gao, R.~Lin, S.~Wang, S.~Bai, S.~Tan, T.~Zhu, T.~Li, T.~Liu, W.~Ge, X.~Deng, X.~Zhou, X.~Ren, X.~Zhang, X.~Wei, X.~Ren, Y.~Fan, Y.~Yao, Y.~Zhang, Y.~Wan, Y.~Chu, Y.~Liu, Z.~Cui, Z.~Zhang, and Z.~Fan.
\newblock Qwen2 technical report.
\newblock \emph{arXiv preprint arXiv:2407.10671}, 2024.

\bibitem[Yang et~al.(2020)Yang, Li, Luo, Wu, and Xu]{yang2020multi}
C.~Yang, Q.~Li, M.~Luo, F.~Wu, and C.~Xu.
\newblock Multi-task learning for image super-resolution with auxiliary tasks.
\newblock In \emph{ECCV}, 2020.

\bibitem[Yang et~al.(2023)Yang, Ping, Liu, Korthikanti, Nie, Huang, Fan, Yu, Lan, Li, et~al.]{yang2023re}
Z.~Yang, W.~Ping, Z.~Liu, V.~Korthikanti, W.~Nie, D.-A. Huang, L.~Fan, Z.~Yu, S.~Lan, B.~Li, et~al.
\newblock Re-vilm: Retrieval-augmented visual language model for zero and few-shot image captioning.
\newblock \emph{arXiv preprint arXiv:2302.04858}, 2023.

\bibitem[Yasunaga et~al.(2022)Yasunaga, Aghajanyan, Shi, James, Leskovec, Liang, Lewis, Zettlemoyer, and Yih]{yasunaga2022retrieval}
M.~Yasunaga, A.~Aghajanyan, W.~Shi, R.~James, J.~Leskovec, P.~Liang, M.~Lewis, L.~Zettlemoyer, and W.-t. Yih.
\newblock Retrieval-augmented multimodal language modeling.
\newblock \emph{arXiv preprint arXiv:2211.12561}, 2022.

\bibitem[Yu et~al.(2024)Yu, Gu, Li, Hu, Kong, Wang, He, Qiao, and Dong]{yu2024scaling}
F.~Yu, J.~Gu, Z.~Li, J.~Hu, X.~Kong, X.~Wang, J.~He, Y.~Qiao, and C.~Dong.
\newblock Scaling up to excellence: Practicing model scaling for photo-realistic image restoration in the wild.
\newblock In \emph{Proceedings of the IEEE/CVF Conference on Computer Vision and Pattern Recognition}, pages 25669--25680, 2024.

\bibitem[Yue et~al.(2024)Yue, Ni, Zhang, Zheng, Liu, Zhang, Stevens, Jiang, Ren, Sun, Wei, Yu, Yuan, Sun, Yin, Zheng, Yang, Liu, Huang, Sun, Su, and Chen]{yue2023mmmu}
X.~Yue, Y.~Ni, K.~Zhang, T.~Zheng, R.~Liu, G.~Zhang, S.~Stevens, D.~Jiang, W.~Ren, Y.~Sun, C.~Wei, B.~Yu, R.~Yuan, R.~Sun, M.~Yin, B.~Zheng, Z.~Yang, Y.~Liu, W.~Huang, H.~Sun, Y.~Su, and W.~Chen.
\newblock Mmmu: A massive multi-discipline multimodal understanding and reasoning benchmark for expert agi.
\newblock In \emph{Proceedings of CVPR}, 2024.

\bibitem[Zamir et~al.(2021)Zamir, Arora, Khan, Hayat, Khan, Yang, and Shao]{zamir2021restormer}
S.~W. Zamir, A.~Arora, S.~H. Khan, M.~Hayat, F.~Khan, M.~Yang, and L.~Shao.
\newblock Restormer: Efficient transformer for high-resolution image restoration.
\newblock In \emph{CVPR}, 2021.

\bibitem[Zhang et~al.(2023{\natexlab{a}})Zhang, Liu, Tan, Le, and Yu]{zhang2023tttllm}
H.~Zhang, Y.~Liu, M.~Tan, Q.~Le, and A.~A. Yu.
\newblock Test-time adaptation for large language models via entropy minimization and alignment tuning.
\newblock \emph{arXiv preprint arXiv:2305.07026}, 2023{\natexlab{a}}.

\bibitem[Zhang et~al.(2025)Zhang, Khayatkhoei, Chhikara, and Ilievski]{zhang2025mllms}
J.~Zhang, M.~Khayatkhoei, P.~Chhikara, and F.~Ilievski.
\newblock Mllms know where to look: Training-free perception of small visual details with multimodal llms.
\newblock \emph{arXiv preprint arXiv:2502.17422}, 2025.

\bibitem[Zhang et~al.(2017)Zhang, Zuo, Chen, Meng, and Zhang]{zhang2017beyond}
K.~Zhang, W.~Zuo, Y.~Chen, D.~Meng, and L.~Zhang.
\newblock Beyond a gaussian denoiser: Residual learning of deep cnn for image denoising.
\newblock \emph{IEEE Transactions on Image Processing}, 26\penalty0 (7):\penalty0 3142--3155, 2017.

\bibitem[Zhang et~al.(2022)Zhang, Sun, Jiang, Yu, Weng, Yuan, Luo, and Kong]{zhang2021bytetrack}
Y.~Zhang, P.~Sun, Y.~Jiang, D.~Yu, C.~Weng, Z.~Yuan, P.~Luo, and T.~Kong.
\newblock Bytetrack: Multi-object tracking by associating every detection box.
\newblock In \emph{ECCV}, 2022.

\bibitem[Zhang et~al.(2023{\natexlab{b}})Zhang, Wang, Xie, Lin, and Wang]{zhang2023vlmttt}
Y.~Zhang, Q.~Wang, P.~Xie, J.~Lin, and X.~E. Wang.
\newblock Test-time adaptation of vision-language models with cross-modal consistency.
\newblock \emph{arXiv preprint arXiv:2310.10039}, 2023{\natexlab{b}}.

\bibitem[Zhu et~al.(2024)Zhu, Tu, Liu, Bovik, and Fan]{zhu2024mwformer}
R.~Zhu, Z.~Tu, J.~Liu, A.~C. Bovik, and Y.~Fan.
\newblock Mwformer: Multi-weather image restoration using degradation-aware transformers.
\newblock \emph{IEEE Transactions on Image Processing}, 2024.

\end{thebibliography}
\bibliographystyle{abbrvnat}


\appendix
\section{Details of the Utilized Restoration Models}
\begin{itemize}[leftmargin=1em,itemsep=0pt]
\vspace{-4pt}
    \item \textbf{NAFNet~\citep{chen2022simple}} is a transformer-based image restoration model that eliminates traditional nonlinear activation functions such as ReLU, GELU, and softmax, replacing them with the simpler SimpleGate mechanism, an element-wise multiplication operation. This design yields a lightweight yet highly effective image restoration framework. The original authors trained NAFNet on the GoPro~\cite{nah2017deep} dataset for image deblurring tasks and on the SIDD~\cite{abdelhamed2018high} dataset for image denoising tasks. In our experiments, we leverage the publicly available pretrained checkpoints to produce restored images for deblurring and denoising. In the subsequent table, we denote images restored by NAFNet as $\mathcal{R_N}$.
    \item \textbf{MWFormer~\citep{zhu2024mwformer}} is a transformer-based model specifically designed for restoring images degraded by various adverse weather conditions. It integrates degradation-aware mechanisms that adaptively handle different weather-induced distortions. In our experiments, we apply MWFormer to restore images affected by snow and fog degradations. In the subsequent table, we denote images restored by MWFormer as $\mathcal{R_M}$.
    \item \textbf{SUPIR~\citep{yu2024scaling}} is a diffusion-based image restoration model that leverages textual guidance and is trained on a dataset comprising 20 million high-resolution, high-quality images. It incorporates low-quality samples into the training process and employs textual prompts to guide the model away from negative visual attributes, thus enhancing visual quality. In our experiments, SUPIR is utilized specifically for image deblurring tasks. In the subsequent table, we denote images restored by SUPIR as $\mathcal{R_S}$.
    \item \textbf{Diffbir~\citep{lin2024diffbir}} introduces a two-staged restoration pipeline that decouples the restoration removal and information regeneration. It utilizes generative diffusion priors to reconstruct high-fidelity images from degraded inputs. In our experiment, we utilize it to restore images degraded by Gaussian noise, denoted as $\mathcal{R_{DB}}$.
    \item \textbf{Daclip~\cite{luo2023controlling}} integrates the CLIP model with a diffusion-based restoration network, IR-SDE~\cite{luo2023image}, to handle various degradation. It also introduced a cross-attention mechanism to injects the content embedding into the diffusion process. In our experiments, Daclip serves as an additional restoration model for images degraded by snow and fog, denoted in the subsequent table as $\mathcal{R_{DC}}$.
\end{itemize}

\section{Additional Experiments and Results}

\subsection{Evaluation of Additional Backbone Models with Degraded Visual Inputs}

In this section, we provide the evaluation results for additional MLLM backbones on MMMU and MathVista with the degraded image input.

\begin{table*}[h]
  \footnotesize
  \setlength{\tabcolsep}{4pt}
  \begin{center}
    \begin{tabular}{l|l|lllll|lllllll}
      \toprule

      \multicolumn{1}{c|}{\multirow{2}{*}{\textbf{Model}}} & \multicolumn{13}{c}{\textbf{MathVista}} \\ \cline{2-14} \multicolumn{1}{c|}{} & ALL& FQA& GPS& MWP& TQA& VQA& ALG &ARI &GEO &LOG &NUM &SCI &STA         \\ \hline

      \multicolumn{1}{l|}{Original} 
      & 45.2
      & 57.6
      & 41.8
      & 31.2
      & 54.4
      & 36.9
      & 42.7
      & 34.6
      & 41.0
      & 8.1
      & 29.2
      & 54.9
      & 61.8
      \\

      Gaussian Noise 
      & 42.0
      & 50.6
      & 39.9
      & 31.2
      & 51.3
      & 34.6
      & 40.2
      & 32.6
      & 40.6
      & 13.5
      & 29.2
      & 54.1
      & 54.8
      \\

      Motion Blur 
      & 32.5
      & 27.9
      & 43.8
      & 15.6
      & 44.9
      & 33.0
      & 40.9
      & 22.1
      & 39.7
      & 10.8
      & 27.8
      & 47.5
      & 26.9
      \\

      Defocus Blur
      & 32.6
      & 26.8
      & 42.3
      & 15.1
      & 48.1
      & 34.6
      & 40.6
      & 22.9
      & 38.5
      & 13.5
      & 26.4
      & 52.5
      & 26.2
      \\

      Snow
      & 42.2
      & 51.3
      & 43.3
      & 29.0
      & 50.6
      & 33.5
      & 42.0
      & 31.2
      & 41.4
      & 5.4
      & 25.7
      & 55.7
      & 56.5
      \\

      Fog
      & 45.2
      & 57.6
      & 41.8
      & 31.2
      & 54.4
      & 36.9
      & 42.7
      & 34.6
      & 41.0
      & 8.1
      & 29.2
      & 54.9
      & 61.8 
      \\

      \bottomrule
    \end{tabular}
  \end{center}
  \caption{Performance of Phi-3-Vision under common image degradations.}
  \label{tab:hallucination-task-mathvista-Phi-3-Vision}
\end{table*}

\begin{table*}[h]
  \footnotesize
  \setlength{\tabcolsep}{4pt}
  \begin{center}
    \begin{tabular}{l|l|lllll|lllllll}
      \toprule

      \multicolumn{1}{c|}{\multirow{2}{*}{\textbf{Model}}} & \multicolumn{13}{c}{\textbf{MathVista}} \\ \cline{2-14} \multicolumn{1}{c|}{} & ALL& FQA& GPS& MWP& TQA& VQA& ALG &ARI &GEO &LOG &NUM &SCI &STA         \\ \hline

      \multicolumn{1}{l|}{Original} 
      & 44.0
      & 48.7
      & 38.5
      & 33.9
      & 55.7
      & 43.6
      & 40.6
      & 37.4
      & 38.1
      & 13.5
      & 35.4
      & 56.6
      & 52.8
      \\

      Gaussian Noise 
      & 41.6
      & 45.0
      & 35.1
      & 34.9
      & 53.2
      & 40.8
      & 37.4
      & 36.0
      & 35.1
      & 13.5
      & 32.6
      & 54.1
      & 50.5
      \\

      Motion Blur 
      & 34.7
      & 33.8
      & 37.5
      & 17.7
      & 48.1
      & 38.5
      & 36.7
      & 26.3
      & 36.0
      & 16.2
      & 29.2
      & 54.9
      & 30.6
      \\

      Defocus Blur
      & 33.8
      & 31.2
      & 37.0
      & 15.1
      & 50.6
      & 38.5
      & 37.0
      & 25.5
      & 33.5
      & 16.2
      & 29.2
      & 56.6
      & 27.6
      \\

      Snow
      & 41.6
      & 43.1
      & 38.9
      & 34.4
      & 53.8
      & 39.1
      & 39.1
      & 35.1
      & 37.7
      & 18.9
      & 31.9
      & 58.2
      & 45.8
      \\

      Fog
      & 44.0
      & 48.7
      & 38.5
      & 33.9
      & 55.7
      & 43.6
      & 40.6
      & 37.4
      & 38.1
      & 13.5
      & 35.4
      & 56.6
      & 52.8 
      \\

      \bottomrule
    \end{tabular}
  \end{center}
  \caption{Performance of Phi-3.5-Vision under common image degradations.}
  \label{tab:hallucination-task-mathvista-Phi-3.5-Vision}
\end{table*}

\begin{table*}[h]
  \footnotesize
  \setlength{\tabcolsep}{5pt}
  \begin{center}
    \begin{tabular}{l|l|llllllllllll}
      \toprule

      \multicolumn{1}{c|}{\multirow{2}{*}{\textbf{Model}}} & \multicolumn{7}{c}{\textbf{MMMU}} \\ 
      \cline{2-8} \multicolumn{1}{c|}{} 
      & Overall 
      & \makecell{Art \&\\Design}
      & Business
      & Science 
      & \makecell{Health \&\\Medicine}
      & \makecell{Humanities \&\\Social Science}
      & \makecell{Tech \&\\Engineering }
      \\ 
      
      \hline

      \multicolumn{1}{l|}{Original} 
      & 38.5
      & 50.0
      & 33.5
      & 30.2
      & 41.0
      & 62.2
      & 34.0
      \\

      Gaussian Noise 
      & 38.0
      & 45.8
      & 32.9
      & 30.0
      & 41.5
      & 61.7
      & 34.0
      
      \\

      Motion Blur 
      & 37.0
      & 47.1
      & 31.9
      & 29.1
      & 41.0
      & 60.5
      & 31.6

      \\

      Defocus Blur
      & 37.1
      & 47.5
      & 31.0
      & 29.0
      & 41.3
      & 60.8
      & 32.3

      \\

      Snow
      & 37.6
      & 45.3
      & 32.6
      & 29.7
      & 41.6
      & 61.4
      & 33.4

      \\

      Fog
      & 38.5
      & 50.0
      & 33.5
      & 30.2
      & 41.0
      & 62.2
      & 34.0

      \\

      \bottomrule
    \end{tabular}
  \end{center}
  \caption{Performance of Phi-3-Vision under common image degradations.}
  \label{tab:hallucination-task-MMMU-Phi-3-Vision}
\end{table*}

\begin{table*}[h]
  \footnotesize
  \setlength{\tabcolsep}{5pt}
  \begin{center}
    \begin{tabular}{l|l|llllllllllll}
      \toprule

      \multicolumn{1}{c|}{\multirow{2}{*}{\textbf{Model}}} & \multicolumn{7}{c}{\textbf{MMMU}} \\ 
      \cline{2-8} \multicolumn{1}{c|}{} 
      & Overall 
      & \makecell{Art \&\\Design}
      & Business
      & Science 
      & \makecell{Health \&\\Medicine}
      & \makecell{Humanities \&\\Social Science}
      & \makecell{Tech \&\\Engineering }
      \\ 
      
      \hline

      \multicolumn{1}{l|}{Original} 
      & 38.5
      & 50.5
      & 32.0
      & 30.8
      & 41.6
      & 62.0
      & 33.6
      \\

      Gaussian Noise 
      & 37.8
      & 47.8
      & 31.7
      & 30.0
      & 40.8
      & 60.6
      & 33.9
      
      \\

      Motion Blur 
      & 37.1
      & 46.7
      & 31.1
      & 29.5
      & 41.1
      & 59.1
      & 32.6

      \\

      Defocus Blur
      & 37.2
      & 46.4
      & 30.3
      & 30.5
      & 39.8
      & 60.4
      & 33.0

      \\

      Snow
      & 37.2
      & 45.1
      & 31.4
      & 30.0
      & 40.6
      & 60.8
      & 33.0

      \\

      Fog
      & 38.5
      & 50.5
      & 32.0
      & 30.8
      & 41.6
      & 62.0
      & 33.6

      \\

      \bottomrule
    \end{tabular}
  \end{center}
  \caption{Performance of Phi-3.5-Vision under common image degradations.}
  \label{tab:hallucination-task-MMMU-Phi-3.5-Vision}
\end{table*}

\subsection{Evaluation with Diffusion-Based Restoration Models}
Table \ref{tab:degrad-eval-restore} presents the complete evaluation results for LLaVA-v1.5-7B~\citep{llava}, LLaVA-v1.6-Mistral-7B~\citep{llavanext}, and Qwen-2.5-VL-3B-instruct~\citep{qwen2.5} under common degradations, as well as transformer-based and diffusion-based model restorations. 
\begin{table*}[htbp]
  \footnotesize
  \setlength{\tabcolsep}{5pt}
  \begin{center}
  \adjustbox{width=1.\linewidth}{
    \begin{tabular}{l|llllllll}
      \toprule
           \textbf{Model}  & MathVista & MMMU & ScienceQA$^T$ & ScienceQA$^I$  & TextVQA & MME$^P$ & MME$^C$       \\ 
      \midrule
      \rowcolor[gray]{0.95}LLaVA-v1.5-7B & 23.3  & 28.7 &  66.02 & 64.85 & 58.18 & 1510.28 & 357.85
      \\

      \rowcolor{gdmg!10}$+$Gaussian Noise & 24.2 & 28.6 & 66.38 & 65.59 & 56.53 & 1431.58 &  341.78
      \\
      $+$Gaussian Noise$+\mathcal{R_N}$ & 17.2 & 31.0 & 65.95 & 64.70 & 56.10 & 1419.18 & 345.00 \\  
      $+$Gaussian Noise$+\mathcal{R_{DB}}$ & 16.6 & 33.1& 65.53 & 63.81 & 50.38 & 1367.62 & 315.36 \\ 
      
      \rowcolor{gdmg!10}$+$Motion Blur & 24.4 & 29.5 & 66.12 & 65.05 & 54.33 & 1454.98 & 361.42 
      \\
        $+$Motion Blur$+\mathcal{R_N}$ & 17.8 & 31.0 & 66.14 & 65.10 & 56.77 & 1473.19 & 317.50 \\
        $+$Motion Blur$+\mathcal{R_S}$ & 23.5 &33.1 & 66.23 & 65.25 & 50.50 & 1465.72 & 323.57 \\ 

      \rowcolor{gdmg!10}$+$Defocus Blur & 24.4 & 29.5 & 66.19 & 65.20 & 54.13 & 1435.93 & 326.07
      \\
      $+$Defocus Blur$+\mathcal{R_N}$ & 16.6 & 31.0 & 66.42 & 65.69 & 53.76 & 1435.69 & 347.86 \\
      $+$Defocus Blur$+\mathcal{R_S}$ & 24.3 &33.1 & 66.09 & 64.95 & 49.67 & 1365.60 & 335.71 \\ 

      \rowcolor{gdmg!10}$+$Snow & 24.1 & 28.2 & 65.36 & 63.46 & 53.24 & 1405.89 & 330.35 
      \\
      $+$Snow$+\mathcal{R_M}$ & 23.1 & 31.0 & 66.05 & 64.85 & 53.85 & 1391.24 & 315.36 & \\  
      $+$Snow$+\mathcal{R_{DC}}$ & 23.5 &33.1 & 65.76 & 64.25 & 53.26 & 1414.90 & 331.07 & \\ 

      \rowcolor{gdmg!10}$+$Fog & 24.8 & 28.6 & 65.39 & 63.51 & 57.13 & 1446.64 & 342.85
      \\
      $+$Fog$+\mathcal{R_M}$ & 23.8 & 31.0 & 65.83 & 64.40 & 56.68 & 1429.61 & 350.71 & \\  
      $+$Fog$+\mathcal{R_{DC}}$ & 23.7 &33.1 & 65.64 & 64.01 & 53.36 & 1412.66 & 322.14 & \\ 
      
      \midrule
      \rowcolor[gray]{0.95}LLaVa-v1.6-Mistral-7B & 26.5 & 34.2 & 76.02 & 71.34 & 63.80 & 1494.22 & 323.92
      \\

      \rowcolor{gdmg!10}$+$Gaussian Noise
      &29.8&33.7&76.44&72.24&59.09&1461.04&307.14\\

      $+$Gaussian Noise$+\mathcal{R_N}$&27.4&34.8&75.95&71.24&58.66&1438.57&303.57\\

      $+$Gaussian Noise$+\mathcal{R_{DB}}$& 25.0 &34.7&75.78&70.90&44.25& 1366.93 & 270.36\\
      \rowcolor{gdmg!10}$+$Motion Blur&25.2&33.6&76.09&71.49&55.13&1454.99&339.64\\
      $+$Motion Blur$+\mathcal{R_N}$&27.8&22.3&76.09&71.54&59.35&1474.13&298.21\\
      $+$Motion Blur$+\mathcal{R_S}$& 23.9 &34.7&75.88&71.1&44.26& 1411.04 & 301.07\\

      \rowcolor{gdmg!10}$+$Defocus Blur
      &25.7& 34.2 &76.4&72.14&51.54&1395.71&275.71\\

      +Defocus Blur$+\mathcal{R_N}$&25.8&22.3&76.63&72.68&51.57&1371.05&285.36&
      \\

      +Defocus Blur$+\mathcal{R_S}$& 25.5 &34.7&76.09&71.54&44.63& 1372.05 & 309.64\\

      \rowcolor{gdmg!10}$+$Snow & 28.7 & 33.7 & 76.00 & 71.29 & 55.39 & 1429.05 & 305.71
      \\
      $+$Snow$+\mathcal{R_M}$&26.5&22.3&76.11&71.59&56.38&1416.26&320
      \\
      $+$Snow$+\mathcal{R_{DC}}$& 26.1 &34.7&75.95&71.24 & 49.55 & 1400.71 & 331.43
      \\
      \rowcolor{gdmg!10}$+$Fog & 30.7 & 34.2 & 76.00 & 71.29 & 62.65 & 1464.83 & 307.14
      \\
      $+$Fog$+\mathcal{R_M}$&25.4&34.8&76.23&71.84&61.16&1477.21&304.29
      \\
      $+$Fog$+\mathcal{R_{DC}}$& 27.3 &34.7&76.26&71.89 & 50.16 & 1404.33 & 315.71
      \\
      \midrule
      \rowcolor[gray]{0.95}Qwen-2.5-VL-3B-instruct & 61.6 & 43.7 & 75.71 & 79.28 & 77.89 & 1515.32 & 615.00
      \\

      \rowcolor{gdmg!10}$+$Gaussian Noise & 57.5 & 42.1 & 73.90 & 75.81 & 65.50 & 1430.75 & 597.14
      \\
      $+$Gaussian Noise$+\mathcal{R_N}$ & 50.2 & 40.2 & 37.92 & 0.00 & 69.28 & 1439.65 & 580.00 \\  
      $+$Gaussian Noise$+\mathcal{R_{DB}}$ & 42.1 &40.6 & 37.99 & 0.00 & 38.28 & 1352.65 & 568.57 \\  

      \rowcolor{gdmg!10}$+$Motion Blur & 56.6 & 41.4 & 74.02 & 75.81 & 61.39 & 1396.84 & 574.64
      \\
      $+$Motion Blur$+\mathcal{R_N}$ & 48.6 & 41.3 & 38.01 & 0.00 & 70.33 & 1502.61 & 557.50 \\  
      $+$Motion Blur$+\mathcal{R_S}$ & 39.2 & 40.6& 38.03 & 0.00 & 38.29 & 1394.83 & 532.50 \\  
 
      \rowcolor{gdmg!10}$+$Defocus Blur & 56.5 & 39.7 & 73.97 & 75.66 & 54.18 & 1346.74 & 555.35
      \\
      $+$Defocus Blur$+\mathcal{R_N}$ & 40.0 & 39.0 & 37.96 & 0.00 & 53.86 & 1399.06 & 530.71 \\  
      $+$Defocus Blur$+\mathcal{R_S}$ & 36.8 &39.8 & 37.77 & 0.00 & 40.36 & 1380.93 & 487.86 \\  

      \rowcolor{gdmg!10}$+$Snow & 52.9 & 41.8 & 73.90 & 75.41 & 64.61 & 1427.12 & 554.28
      \\
      $+$Snow$+\mathcal{R_M}$ & 45.6 & 40.5 & 37.87 & 0.00 & 65.35 & 1443.55 & 581.43 \\  
      $+$Snow$+\mathcal{R_{DC}}$ & 45.5 &41.8 & 37.92 & 0.00 & 48.95 & 1286.25 & 541.07  \\ 

      \rowcolor{gdmg!10}$+$Fog & 61.6 & 43.2 & 74.89 & 77.54 & 73.69 & 1499.79 & 605.71
      \\
      $+$Fog$+\mathcal{R_M}$ & 48.8 & 42.1 & 37.82 & 0.00 & 73.62 & 1478.08 & 626.43 \\  
      $+$Fog$+\mathcal{R_{DC}}$ & 48.2 & 43.9& 37.87 & 0.00 & 51.83 & 1396.19 & 600.36 \\ 
      
      \midrule
      \rowcolor[gray]{0.95}VLM-R1 & 63.1 & 44.2 & 69.18 & 75.36 & 76.58 & 1523.19 & 658.21
      \\

      \rowcolor{gdmg!10}$+$Gaussian Noise & 60.7 & 43.0 & 67.58 & 71.99 & 64.96 & 1388.92 & 636.42
      \\
      $+$Gaussian Noise$+\mathcal{R_N}$ & 53.3 & 42.6 & 32.92 & 0.00 & 68.70 & 1436.39 & 616.07 &\\  
      $+$Gaussian Noise$+\mathcal{R_{DB}}$ & 42.0 &40.2 & 32.92 & 0.00 & 37.72 &1375.77 & 577.86&\\  

      \rowcolor{gdmg!10}$+$Motion Blur & 57.5 & 42.7 & 67.81 & 72.48 & 60.90 & 1487.96 & 586.79 
      \\
      $+$Motion Blur$+\mathcal{R_N}$ & 49.2& 42.3 & 32.92 & 0.00 & 69.85 & 1487.96&586.79 &\\  
      $+$Motion Blur$+\mathcal{R_S}$ & 38.9 &39.9 & 32.92 & 0.00 & 37.91 &1411.46 &509.29 &\\  
 
      \rowcolor{gdmg!10}$+$Defocus Blur & 59.0 & 41.6 & 67.13 & 71.05 & 53.84 & 1396.31 & 545.35 
      \\
      $+$Defocus Blur$+\mathcal{R_N}$ &41.2 & 40.5 & 32.92 & 0.00 & 52.89 & 1390.23 & 510.00 &\\  
      $+$Defocus Blur$+\mathcal{R_S}$ &38.9 &39.8 & 32.92 & 0.00 & 40.16 &1379.95 & 483.57&\\  

      \rowcolor{gdmg!10}$+$Snow & 53.3 & 42.3 & 67.06 & 70.90 & 62.99 & 1360.11 & 571.42
      \\
      $+$Snow$+\mathcal{R_M}$ & 47.6 & 41.9 & 32.92 & 0.00 & 64.27 & 1416.31& 597.5&\\  
      $+$Snow$+\mathcal{R_{DC}}$ & 47.2 & 41.8& 32.92 & 0.00 & 48.66 & 1292.01&548.57&\\  

      \rowcolor{gdmg!10}$+$Fog & 63.2 & 44.2 & 68.17 & 73.23 & 72.55 & 1487.00 & 612.85
      \\
      $+$Fog$+\mathcal{R_M}$ & 52.3& 43.4 & 32.92 & 0.00 & 72.68 & 1478.15&618.93 &\\  
      $+$Fog$+\mathcal{R_{DC}}$ &50.2 &44.1 & 32.92 & 0.00 & 51.53 &1355.76 & 575.00&\\  

      \bottomrule
    \end{tabular}}
  \end{center}
  \caption{Performance of MLLMs across common degradations and restorations.}
  \label{tab:degrad-eval-restore}
\end{table*}


\section{Examples of the Input Images with Degradations and Restorations}
\begin{figure}
    \centering
    \includegraphics[width=1.\linewidth]{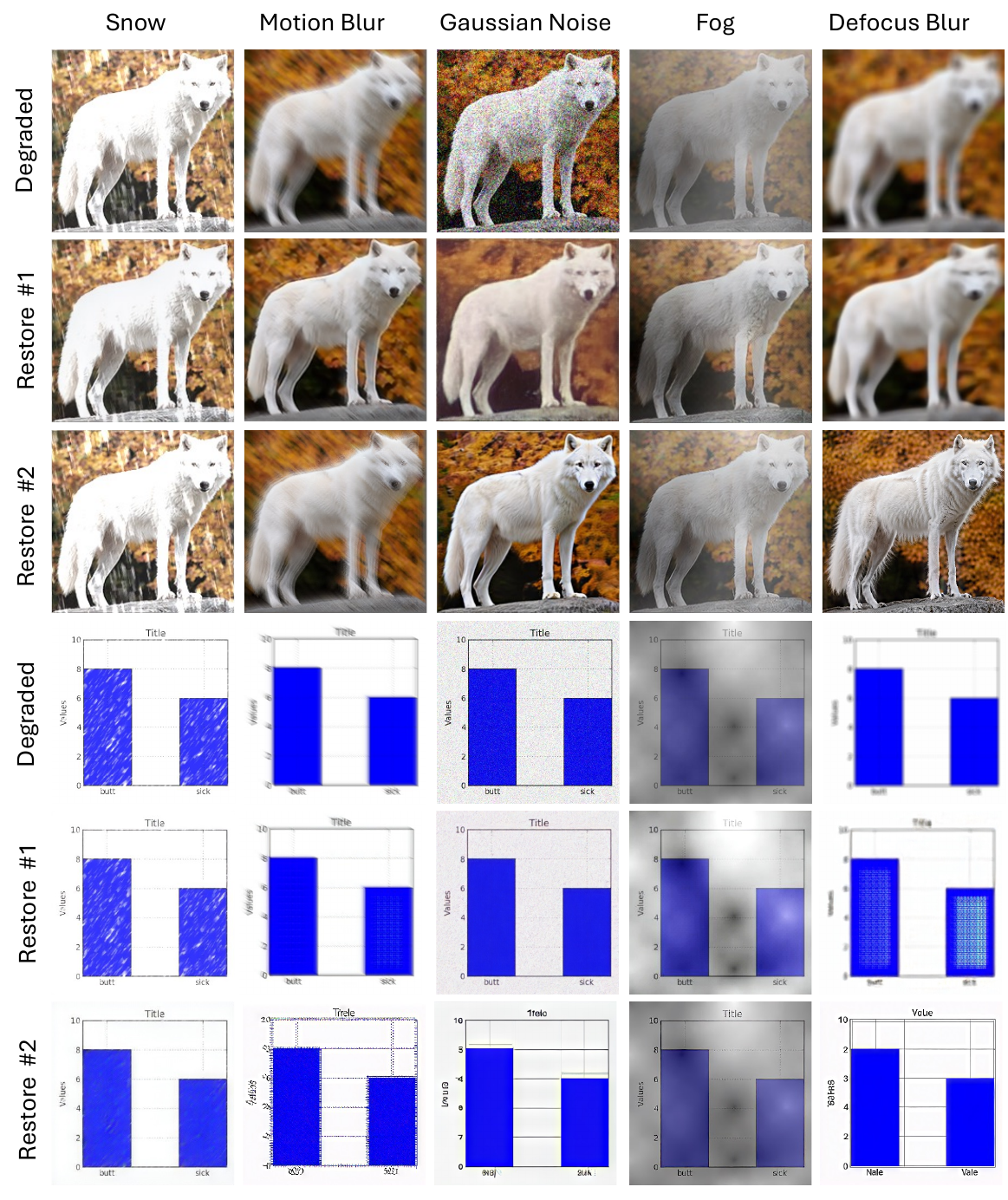}
    \caption{Examples of the images with degradations and restorations.}
    \label{fig:img-ddegrad}
\end{figure}

Examples of the input images with degradations and restorations are provided in Figure \ref{fig:img-ddegrad}.

\section{Broader Impacts}
\label{sec:impact}
The social impacts of our work are mainly in three folds: 
\begin{enumerate}
    \item \textbf{Accessibility:} This work provides a lightweight and training-free framework to improve the performance of Multimodal Large Language Models (MLLMs) across a wide range of visual inputs. By adapting input images on-the-fly without requiring high-end GPUs, external models, or curated datasets, VQ-TTT enables broader and more inclusive use of MLLMs in real-world scenarios, especially where image quality is suboptimal or device capabilities are limited.
    
    \item \textbf{Applications:} Our findings challenge the conventional assumption that higher perceptual fidelity always leads to better machine understanding, with implications for education, accessibility technologies, content moderation, and human-AI collaboration. VQ-TTT can improve MLLM reliability in diverse environments such as mobile photography, remote sensing, telemedicine, and surveillance—where visual degradation is common and high accuracy is critical.
    
    \item \textbf{Open Access:} We are committed to open-sourcing our code and models to encourage further exploration of visual quality modulation in MLLMs. By doing so, we aim to support transparency, reproducibility, and collective progress in building adaptive, robust vision-language systems that serve a wide range of users and use cases.
\end{enumerate}
We do not foresee serious negative societal impacts from this work. Our method operates within the input space and enhances model alignment without altering foundational model behavior or generating misleading content. Instead, it highlights the importance of understanding and adapting to AI models' perceptual biases for more reliable deployment.

\section{Limitations}
\label{sec:limitation}
While VQ-TTT is designed to be lightweight and efficient, its simplicity also introduces limitations. To enable fast adaptation across diverse inputs, we intentionally restrict the number of tunable parameters through low-rank modulation and shallow-layer LoRA updates. However, this compact design may limit the model’s capacity to handle complex or highly nonlinear degradation patterns. In such cases, generalization performance may degrade, especially when the distortions fall outside the scope of the training-time assumptions.

Additionally, our current evaluation is limited in scope. While we benchmark across several representative MLLMs and datasets, the coverage is not exhaustive. The number of tested models, vision-language tasks, and degradation types remains relatively narrow. Future work should investigate broader model families, more diverse real-world datasets, and a wider spectrum of degradation modalities (e.g., motion blur, occlusion, sensor noise) to more comprehensively assess the robustness and applicability of VQ-TTT.

\end{document}